\documentclass{article} % For LaTeX2e
\usepackage{iclr2027_conference,times}

\usepackage{amsmath,amsfonts,bm}

\def\eqref#1{equation~\ref{#1}}
\def\1{\bm{1}}

\DeclareMathAlphabet{\mathsfit}{\encodingdefault}{\sfdefault}{m}{sl}
\SetMathAlphabet{\mathsfit}{bold}{\encodingdefault}{\sfdefault}{bx}{n}

\usepackage{hyperref}
\usepackage{url}
\usepackage{titletoc}
\usepackage{xspace}
\usepackage{graphicx}
\usepackage{algorithm,algpseudocode}
\usepackage{cleveref}
\usepackage{booktabs}
\usepackage{tabularx}
\newcommand\DoToC{%
  \startcontents
  \printcontents{}{1}{\hrulefill\vskip0pt}
  \vskip0pt \noindent\hrulefill
  }

\definecolor{cornflowerblue}{rgb}{0.39, 0.58, 0.93}
\hypersetup{
    colorlinks=true,
    linkcolor=cornflowerblue,
    filecolor=magenta,      
    urlcolor=cornflowerblue,
    citecolor=cornflowerblue,
    pdftitle={LoL},
    pdfpagemode=FullScreen,
}

\newcommand{\name}{LoL\xspace}
\newcommand{\fullname}{Library of Layers\xspace}

\title{Library of Layers (LoL): Adaptively Ordering Transformer Layers at Runtime}

\author{%
  \textbf{Abdellah Rahmani}$^{1*}$,\;
  \textbf{Sean McLeish}$^{1,2*}$,\;
  \textbf{Ang Li}$^{3}$,\;
  \\ \textbf{Sanae Lotfi}$^{3}$,\;
  \textbf{Pascal Frossard}$^{1}$,\;
  \textbf{Micah Goldblum}$^{4}$\\[2pt]
  $^{1}$EPFL, $^{2}$University of Maryland,
  $^{3}$New York University, $^{4}$Columbia University
}
\iclrpreprint % Uncomment for camera-ready version, but NOT for submission.
\begin{document}

\maketitle
\ificlrfinal
{\renewcommand{\thefootnote}{}%
\footnotetext{$^{*}$Equal contribution. Correspondence to:
\href{mailto:abdellah.rahmani@epfl.ch}{abdellah.rahmani@epfl.ch},
\href{mailto:smcleish@umd.edu}{smcleish@umd.edu}\\
\hspace*{2.2em}Code: \href{https://github.com/arahmani1/library-of-layers}{github.com/arahmani1/library-of-layers}}}
\fi

\begin{abstract}
Computer programmers chain together the same reusable computational primitives, for example, conditional statements or loops, but in different orders to implement different algorithms.  In contrast, existing language model architectures always execute their layers in the same order.  We introduce \fullname (\name), an architecture that dynamically orders its layers during inference to solve the problem at hand.  \name has a single hyperparameter,  the size of its global library of
transformer blocks, that allows it to transition seamlessly between parameter reuse and sparsity, generalizing the parameter-efficiency and FLOPs-efficiency tradeoffs of dense, depth-recurrent, and mixture of experts architectures in one unified family.  We show that \name achieves competitive performance with all three above baselines when pretrained on text data.  On synthetic sudoku and maze data, we observe that the model employs distinct layer orderings to implement different solution strategies, and we can intervene on routing to force a model to apply a specific solution strategy to a problem instance.  In a multilingual natural language setting, we find that \name chooses different layer order patterns based on language and script.
Similarly, we train \name on safety data and fit probes to its router outputs, finding that these probes distinguish safe from unsafe text with high AUROC.
\name provides a path toward a future in which models dynamically compose their own computational graph, enabling more efficient and controllable reasoning.
\end{abstract}

\section{Introduction}
Many distinct algorithms are built from the same set of computational primitives, composed in different orders to implement different behaviors.
For example, a database query engine can combine operations such as filtering, joining, sorting, and aggregation in different sequences depending on the query being executed. Modular programming lets the same reusable components be dynamically composed for many purposes within a single codebase. Transformer language models lack this flexibility; their layers are executed in a fixed, predefined order for every input. This rigidity may make these models less efficient, since useful computational primitives may need to be represented repeatedly at different depths.

Prior work has focused on increasing the sparsity of models using mixture of experts (MoE) \citep{shazeer2017outrageously,fedus2022switch} or reusing layers multiple times during the forward pass in looped models \citep{dehghani2018universal,lan2019albert,geiping2026scaling}.
Other work has investigated skipping layers dynamically during the forward pass \citep{elhoushi2024layerskip,raposo2024mixture,bae2025mixture}.
However, all of these approaches still enforce the order in which the layers of the model must be executed.

\begin{figure}[t!]
    \centering
    \includegraphics[width=1\linewidth]{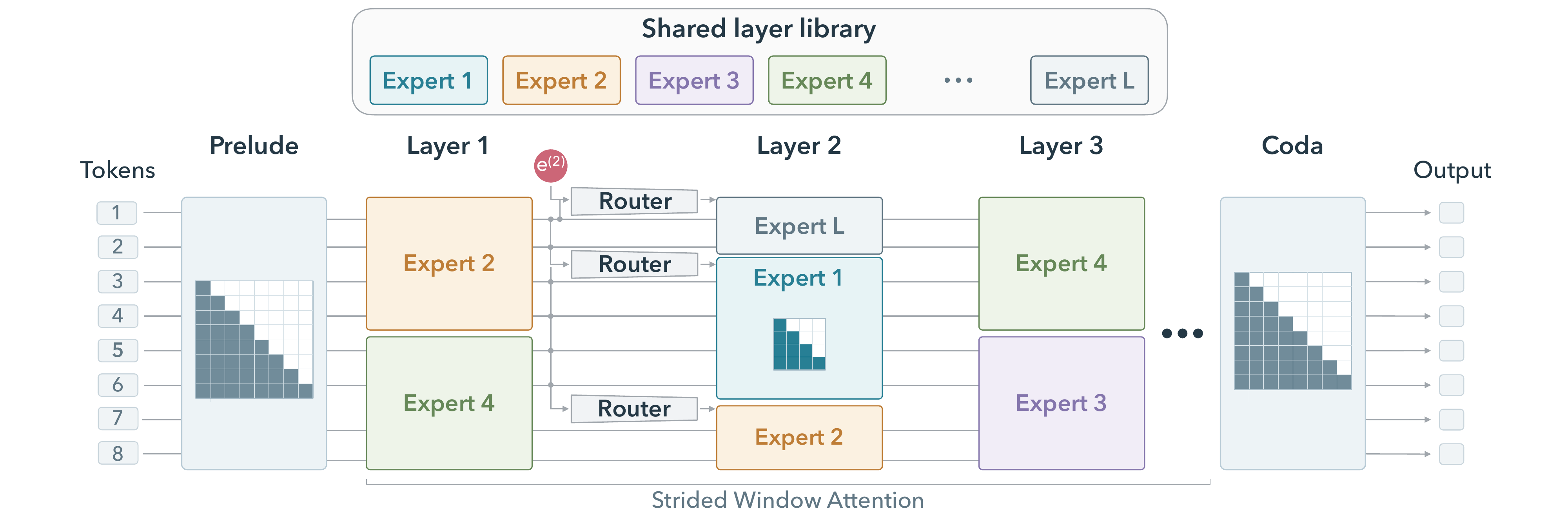}
    \caption{\textbf{We introduce \fullname (\name), a novel architecture where the next executed transformer layer is chosen at runtime.} We use \textit{strided window attention} within the routed core to allow information to flow through the sequence dimension while using the same attention and MLP parameters for all tokens within each window.}
    \label{fig:cover}
\end{figure}

In this paper, we introduce \textit{\fullname} (\name), an architecture in which a small router, trained end-to-end without explicit supervision, dynamically selects the next layer from a global library of transformer blocks.
We note this architecture encompasses the two properties explored by prior work.
A small library of layers forces layer reuse, whereas a large library of layers allows many layers to remain inactive on a given forward pass. By changing only the library size, LoL therefore spans a continuum from highly reused, dense computation to highly sparse computation.
We visualize our novel architecture in \Cref{fig:cover}.

Dynamic layer selection introduces a challenge.  We want to allow different token positions the flexibility to select different layer orderings, but this means that a vector at one token position extracted by a selected attention layer might attend to vectors at previous token positions extracted by different selected attention layers.  We observe that this mismatch makes training unstable and harms performance.  At the same time, if we chose a single layer ordering for the entire context, we would risk either breaking causality or making layer decisions with too little information (only the prompt).

We therefore introduce a windowed attention masking strategy that simultaneously allows different windows to adaptively select layer orderings while also preventing the above mismatch problem and preserving causality.
We partition the context into windows, route each window using only information available from earlier windows, and shift the window boundaries across routed depths so that information can propagate between neighboring windows, expanding the field of view with depth.
% We note that these boundaries depend only on the routed depth: they are fixed in advance and are independent of both the input sequence and the layer selected by the router.
% This makes the scheme easy to set, with only a window size and a stride to choose, and means the model always knows where the next window begins, in training as well as in generation.
% It also gives the router full flexibility, since any expert can be placed at any depth without changing how the sequence is split.

We use synthetic and natural-language experiments to study and interpret our models’ learned layer orderings. On sudoku and maze tasks, \name learns distinct layer-ordering patterns for different solution strategies; forcing a strategy's learned routing path at inference time causes the model to follow that strategy.
This phenomenon is especially clear in mazes, where all algorithms we consider use the same output syntax and differ only in the order in which they explore cells.
We find that fixing layer ordering controls the cell exploration order itself.
In multilingual language modeling, different languages use distinct execution paths, with languages that share scripts exhibiting similar routing patterns.
We also train \name on a mix of safe and unsafe data, finding that a ridge regression trained on routing features distinguishes these categories on held-out data with high AUROC.
On general web text, we vary the library size to induce parameter reuse regimes with a FLOPs-to-parameters ratio comparable to that of looped transformers and also to induce sparse regimes with a FLOPs-to-parameters ratio comparable to MoE models, finding that \name remains competitive across this range.
Finally, at 1B parameters on math data, \name is competitive with conventional dense transformers in terms of loss and benchmark accuracy in a FLOPs-matched setting.

These results suggest that layer ordering can become a learned part of the computation itself rather than a fixed architectural choice.
\name unifies behaviors associated with seemingly distinct dense, depth-recurrent, and MoE models by varying a single knob, the size of the layer library, while exposing the model's internal execution path.
Because routing patterns can be extracted and intervened on, \name also provides a simple mechanism for studying and controlling latent computation.
%Post-hoc interpretability for latent reasoning models has so far had varying degrees of success \citep{dilgren2026latent,lu2025latent}.

Our contributions are summarized as follows:
\begin{itemize}
  \item We propose \textit{\fullname} (\name), a novel architecture in which a router selects blocks to apply from a global library of transformer layers, making layer selection, ordering, and the number of times each layer is used conditional on the input.
  \item We use maze, sudoku, multilingual, and safety training data to empirically show the \name router learns input-adaptive patterns from data. Moreover, these router patterns can then be used to control the model at test time by forcing the layer execution ordering.
  \item We demonstrate that \name models perform competitively across a spectrum of sparsity levels, from low-sparsity looped models that repeatedly reuse the same layers to high-sparsity mixture of experts models.
\end{itemize}

\section{Related Work}
Prior architectures reuse parameters across depths or select a subset of parameters at each depth.
\fullname spans both regimes by varying the size of its shared library of transformer blocks.

\begin{figure}[t!]
    \centering
    \includegraphics[width=1\linewidth]{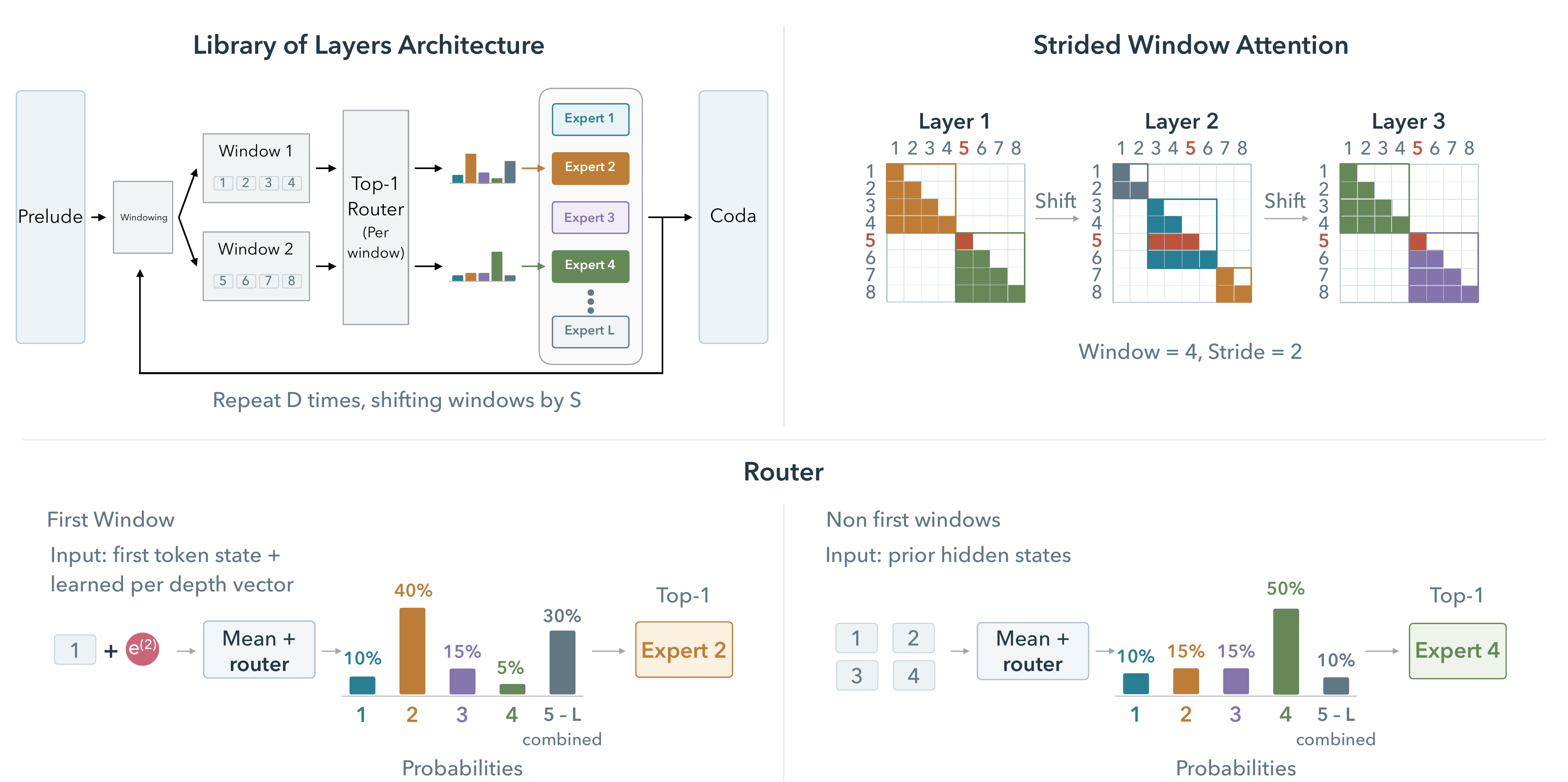}
    \caption{\textbf{\fullname uses strided window attention and a top-1 router to allow for efficient training.}
\name is made up of a prelude, a routed core, and a coda.
The layers in the prelude and coda can attend over the whole sequence.
In the core, the sequence is split into windows of $W$ tokens, and at every routed depth a router selects one layer from the shared library per window (top-left).
Attention is restricted to each window, but the window boundaries shift by $S$ tokens between successive routed depths, allowing information to flow across windows (Strided Window Attention, top-right).
Window $0$ has no history, so its router uses the first token and a learned vector $e^{(\ell)}$ for routed depth $\ell$ (bottom-left).
For window $i>0$, the router averages the mean hidden states of earlier windows $0,\dots,i-1$ and selects one layer (bottom-right).
With $W=4$ and $S=2$, Layers 1 and 3 share the same boundaries while Layer 2 is offset, so information can pass between neighboring windows.
Token 5 is highlighted in red to show this: in Layer 1 it is the first token in a new window, while in Layer 2 its window includes tokens 3 and 4.
}
    \label{fig:explainer}
\end{figure}

\textbf{Parameter Sharing.}
Looped transformers, which repeatedly reuse the same transformer blocks, are efficient and can be made Turing complete \citep{dehghani2018universal,lan2019albert,giannou2023looped,csordas2024moeut,saunshi2025reasoning,mcleish2025teaching,geiping2026scaling,prairie2026parcae,huang2026towards}.
Looped transformers have also been studied in the context of synthetic algorithmic tasks, showing they can outperform fixed-depth counterparts in both learning efficiency and accuracy \citep{giannou2023looped,yang2023looped,mcleish2024transformers,fan2024looped}.
Recurrence in other architectural regimes such as graph and convolutional neural networks has also shown promising results \citep{schwarzschild2021can,bansal2022end,ibarz2022generalist,schwarzschild2023deep,bear2024rethinking}.
Parameter sharing has been explored for many components of the modern transformer architecture, for example, QKV weights and KV caches across layers \citep{sun2024you,kayyam2026transformers}.
We build on these works by incorporating architectural choices from looped transformers, such as only using recurrence in the middle layers of the model; see \Cref{fig:explainer} (top-left).

Most similar to our work is that of \citet{gong2024mixture}. However, \citet{gong2024mixture} pretrain their model before adding layer reuse and do not dynamically choose each layer in the final model.
\citet{gong2024mixture} use top-k routing with k greater than one, unlike our top-1 routing. They avoid attention space mismatches by averaging the raw weights of the selected experts in the attention computation and combine the output of the FFN via a weighted mean based on the router output weighting.

\textbf{Routing mechanisms.}
Mixture of experts (MoE) layers \citep{shazeer2017outrageously} have become the norm for large transformer-based models \citep{openai2025gptoss120bgptoss20bmodel,xu2026deepseek}, often implemented with top-k token-based routing according to the current hidden state.
\citet{huang2026unipool} extend this to share a global pool of experts so that all layers can select any expert.
Others add a skip option to the router, allowing tokens to dynamically skip layers \citep{raposo2024mixture,bae2025mixture}.
\citet{ternovtsii2026mixture} route through multiple smaller transformer layers instead of only routing the MLP, as done in a conventional MoE.
\name goes beyond choosing which computation to apply at a fixed position in the network: each window selects a complete transformer block from a library shared across all routed depths, making both block ordering and reuse input-dependent.

\section{How to Learn the Order of a Forward Pass}\label{sec:setup}

\subsection{The Library}
The architecture is built from decoder-only transformer blocks \citep{vaswani2017attention}, organized into a prelude, a routed core, and a coda.
Following prior work on looped transformers \citep{saunshi2025reasoning,geiping2026scaling}, we use adaptive layer selection only in the middle of our \name models.

The \textit{prelude} $P$ and \textit{coda} $C$ are fixed-order stacks of decoder layers with full causal attention, at the beginning and end of the architecture, respectively.
They surround the \textit{routed core}, which applies blocks selected from a library $\{E_1,\ldots,E_L\}$.
Library size $L$ counts the distinct blocks stored, whereas core depth $D$ counts the block applications executed.
We call each library block an \textit{expert} and each application a \textit{routed depth}.
For example, with $L=3$ and $D=3$, the sequence $E_1 \to E_3 \to E_1$ repeats one expert and leaves another unused; we call this sequence a \textit{routing path}.
Different windows of tokens can follow different routing paths, as described below.
We visualize our forward pass in Figures \ref{fig:cover} and \ref{fig:explainer}, and provide pseudocode in \Cref{alg:forward-pass} in the appendix.

\subsection{Routing}
Routing must preserve causality, and the router may only use information that will also be available during generation. The expert selected for a token cannot depend on later tokens since that would violate causality.
One could select an expert for the whole sequence using only the first token to preserve causality, but this would prevent the layer ordering from adapting to information revealed later in the sequence.
Alternatively, routing each token independently could assign neighboring tokens to different experts, each with its own query, key, and value projections.  In this case, queries and keys computed on different attention heads would be compared, and we find empirically that this setup leads to training instability and poor performance. 
To overcome both problems, we chunk the sequence at each depth and route separately for each chunk.  We then stride the chunk boundaries every depth to allow information to flow between adjacent chunks, similarly to how the field of view under sliding window attention increases with depth.
All tokens in a window use the same expert with causal attention restricted to that window, preventing the Q-K mismatch.
% The router selects this expert using hidden states from earlier windows.

\subsubsection{Strided Window Attention}
Specifically, at routed depth $\ell$, let $W$ be the window size and $S$ the stride between depths.
We shift the window boundaries by $\delta_\ell=(\ell-1)S \bmod W$ positions.
% : the shift grows by $S$ at each depth and wraps around once it reaches $W$.
For example, with $n=8$, $W=4$ and $S=2$, depth $1$ uses the windows $\{0,\ldots,3\}$ and $\{4,\ldots,7\}$, depth $2$ uses $\{0,1\}$, $\{2,\ldots,5\}$ and $\{6,7\}$, and depth $3$ goes back to the first split (\Cref{fig:explainer}, top-right).
Formally, position $i \in \{0,\ldots,n-1\}$ belongs to window $I_k$ when $w^{(\ell)}_i = k$, where:
\(
w^{(\ell)}_i
= \left\lfloor
    \frac{i+\delta_\ell}{W}
  \right\rfloor ,
\)
which gives $K_\ell$ windows at depth $\ell$.
We visualize this in \Cref{fig:explainer} (top-right).

All tokens in a window share one selected expert and router weight.
Within a window, each token may attend only to itself and earlier tokens in that window.
Striding the boundaries between depths causes windows at successive depths to overlap, allowing information to flow between adjoining windows while preserving causality.
In the prelude and coda, we use full-sequence causal attention.

\subsubsection{The Router}
When selecting an expert for each window $k>0$, the router receives information only from prior windows $0\leq j<k$ to preserve causality.
Precisely, the router is a linear layer and the input is the mean over windows of the mean hidden state per window.
For the first window $I_0$, there are no prior windows to use.
We instead input the mean of a learned depth-specific vector $\mathbf{e}^{(\ell)}$ and the first token's hidden state to the router.
We show this in \Cref{fig:explainer} (bottom).

To train the routed core, we use loss-free load balancing \citep{wang2024auxiliary,liu2024deepseek}.
The selected expert's residual update is scaled by its router probability, allowing gradients to flow to the router.
Following \citet{shazeer2017outrageously}, we add Gaussian noise to the router logits before the softmax to encourage exploration during training.
Compared with routing per token in MoE models, routing per window provides a coarser training signal: gradient contributions from all tokens in a window are aggregated through a single shared routing decision, rather than separate decisions for each token.
We use FiLM \citep{perez2018film} to add a learned depth-dependent scale and bias after each normalization in the transformer block.
We describe our implementation of FiLM in Appendix \ref{app-sec:training-details} for completeness.

\subsection{Increasing Depth During Training}
The training cost of \name is mainly controlled by the routed core depth (\(D\)). 
Small \(D\) would make training cheap by reducing FLOPs per parameter but would lead to a shallower model with poorer performance.
Therefore, we begin training with a shallow routed core and gradually increase \(D\) over the course of training.
This procedure saves compute by performing much of training with a shallower model, but ultimately constructs a deep and expressive model for deployment.
This growing procedure allows the routers to explore experts quickly and cheaply during early training, as the depth of the model is smaller.
This is similar to increasing the number of loops of a recurrent model over training \citep{mcleish2025teaching} or progressive stacking \citep{gong2019efficient} when training vanilla transformers. We refer to this technique as growing.

Additional architecture and training details are provided in Appendix \ref{app-sec:training-details}.
We provide ablations for our design decisions in Appendix \ref{app-sec:arch-abl}.

\section{Experiments}\label{sec:experiments}
We conduct an analysis of \name's capabilities using a range of synthetic and real-world data.
We begin by using multilingual text, safety text, and synthetic tasks to trace what \name learns during training, and we find that the path the model uses is often interpretable in these regimes.
Subsequently, we train \name on a mixture of math data, showing \name can learn to perform language modeling and mathematical reasoning.
Finally, we vary the sparsity of \name models while training on web text and find \name is competitive with widely used architectures at a wide range of sparsities.

For our \(360M\) parameter models, we use \(4\) layers in the prelude, \(4\) layers in the coda, and a depth of \(16\) in the routed core.

\subsection{What Routing Paths Encode and Control}\label{subsec:exp-router}
We begin by studying the router behavior in four training tasks: multilingual text, safety text, sudoku, and mazes.
We first examine whether routing features distinguish languages and safety labels, then intervene on routing paths to control the strategies used to solve sudoku puzzles and mazes.

\begin{figure}[t!]
    \centering
    \includegraphics[width=\linewidth]{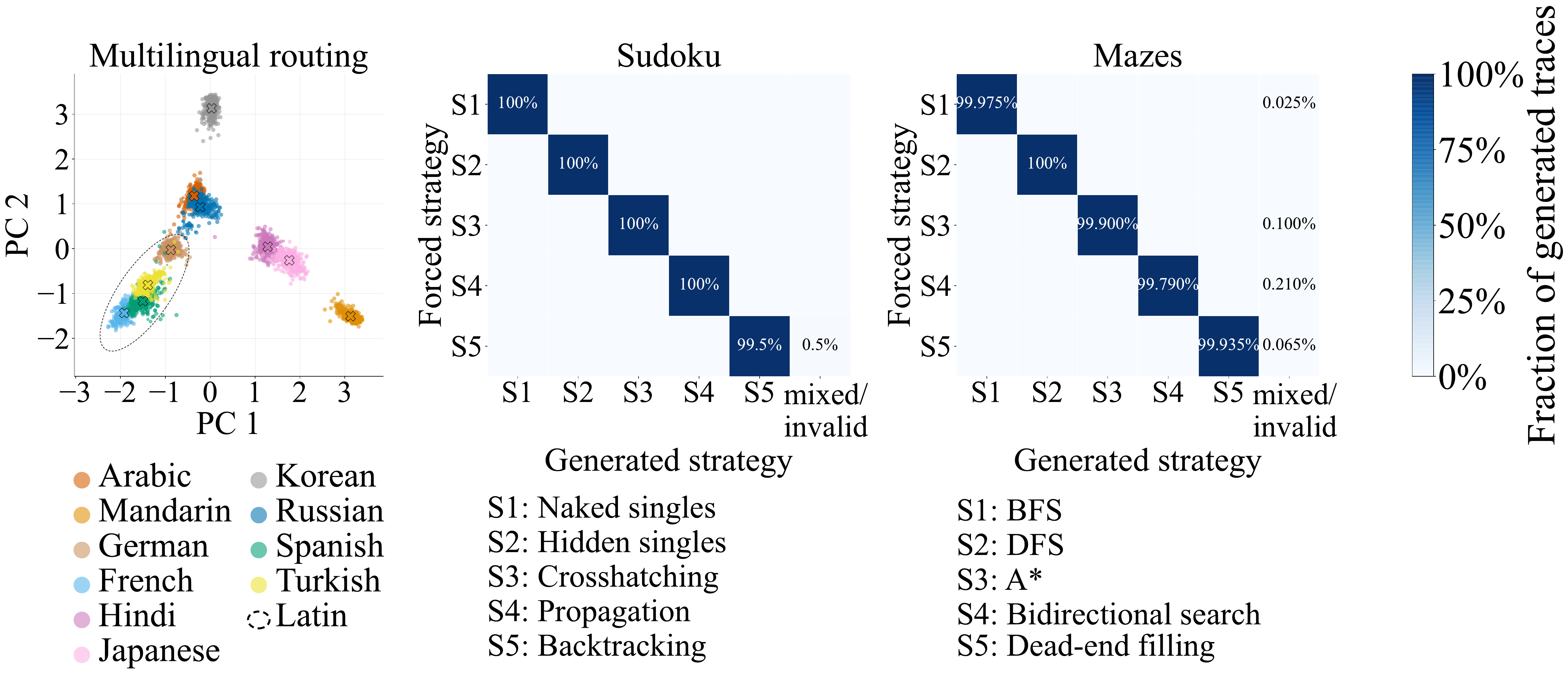}
    \caption{\textbf{\fullname learns distinct paths for different languages and distinct strategies when solving sudoku puzzles and mazes.}
    Left: PCA projection of the routing paths of a \(360M\) parameter \name model trained on an equal mixture of \(10\) languages. We see the paths taken are distinct per language and Latin-script languages use similar paths.
    Middle and Right: We train \(360M\) parameter \name models to solve sudoku (middle) and mazes (right), using \(5\) different strategies. We then force each strategy's mode path at inference time and see the models use the correct strategy (algorithm) with close to \(100\%\) accuracy.
    }
    \label{fig:router}
\end{figure}

\subsubsection{Multilingual}
First, we train \(360M\) parameter models on an equal mix of \(10\) different languages from FineWeb-2 \citep{penedo2025fineweb2} for \(10\) billion tokens.
On a held-out validation set, for each sampled window, we encode the router’s selected expert at every routed depth as a one-hot vector.
We then concatenate these one-hot vectors and project the resulting vector into two dimensions using PCA.
In \Cref{fig:router} (left), we see that the routing paths are reasonably distinct per language and the Latin-script languages all share similar paths.
Each point in the plot represents one window and its sequence of selected experts.
In Appendix \Cref{app-fig:multilang}, we see \name gradually learns to distinguish between languages and route them through different computational paths by dynamically reordering the experts in its library over the course of training.  Note that the same layers are used for various languages but just executed in different orders.

\subsubsection{Safety}\label{subsubsec:safety}
We now ask whether routing probabilities encode safety-relevant information in natural text.
Using the harmfulness scores assigned to FineWeb sequences by \citet{maini2026safety}, we treat score-0 sequences as safe and score-4 and score-5 sequences as unsafe.
We train models with the same architecture as our multilingual models for \(10B\) tokens on an equal mixture of safe and unsafe text, without providing the harmfulness labels during training.

We train a linear ridge regression to predict the sequence's harmfulness label from a window's router probabilities across all routed depths.
We use a held out set of \(50M\) tokens for training the linear regression, split into \(40M\) for development and \(10M\) for testing.
The probe achieves a held-out test set AUROC of \(0.92\), showing that router probabilities contain linearly accessible information about the dataset's harmfulness labels.

Because each label applies to an entire sequence, an individual window may not contain the content responsible for that label.
We therefore also train a sequence-level ridge probe using router probabilities averaged across windows.
This probe achieves a test set AUROC of \(0.95\).
Appendix \ref{app-sec:safety} shows that simpler readouts recover much of the predictive signal in routing features, while label shuffling controls yield chance performance.

\subsubsection{Sudoku}
Next, we analyze forcing routing paths to control the sudoku solving strategy.
We train the model with strategy information available only to the router, then extract a representative routing path for each strategy.
At inference, we remove the strategy information and force the model to follow the corresponding path.

We construct our dataset including five sudoku solving strategies, each with its own step syntax, fully described in \Cref{app-tab:sudoku-data} in the appendix.
Our data is formatted as \texttt{Q : <Input Grid> [Sx] R : <Solving Steps> A : <Final Grid> <eos>}, where \texttt{[Sx]}, with \texttt{x}\(\in\{1,2,3,4,5\}\), identifies the solving strategy used to generate the reasoning trace.
As the first window for all inputs is just the sudoku grid, we need the \texttt{[Sx]} identifier to tell the model which algorithm to use at test time.
The strategy description tokens are masked from all parts of the network but the router.
We train for approximately \(10B\) tokens, computing the loss over only the solving steps and final grid, masking the question and strategy description.

For each strategy, we find the \textit{mode path} by running the trained model with teacher forcing on held-out validation puzzles containing the strategy description.
For each window and routed depth, we select the expert with the highest mean router probability across these puzzles and fix its residual weight to that mean.
Here, mode refers to the highest mean probability, not the most frequent discrete choice; the path fixes both expert choices and residual weights.
We report both the mode paths and how consistently the router agrees with the mode path at each depth in Appendix Figures \ref{app-fig:sudoku_forced_path} and \ref{app-fig:sudoku_agree}, respectively.

We then test \name on new sudoku grids, replacing the strategy description with a neutral filler, \(\texttt{[...]}\), and forcing each window to follow the corresponding \textit{mode path} for the requested strategy.
This tests whether fixing the path is enough to make the model follow that strategy.
A trajectory is marked correct only if every step is legal, follows the requested strategy’s syntax and rules, and leads to the correct solution.

In \Cref{fig:router} (middle), we plot the confusion matrix where we force the mode path for each strategy.
We see that the model uses the expected strategy with close to 100\% accuracy when the mode path is forced for that strategy.
Appendix \Cref{app-fig:sudoku_results} also shows that \name solves sudoku puzzles with close to 100\% accuracy when given the strategy token \texttt{[Sx]} instead of forcing a routing path.

\begin{figure}[t!]
    \centering
    \includegraphics[width=\linewidth]{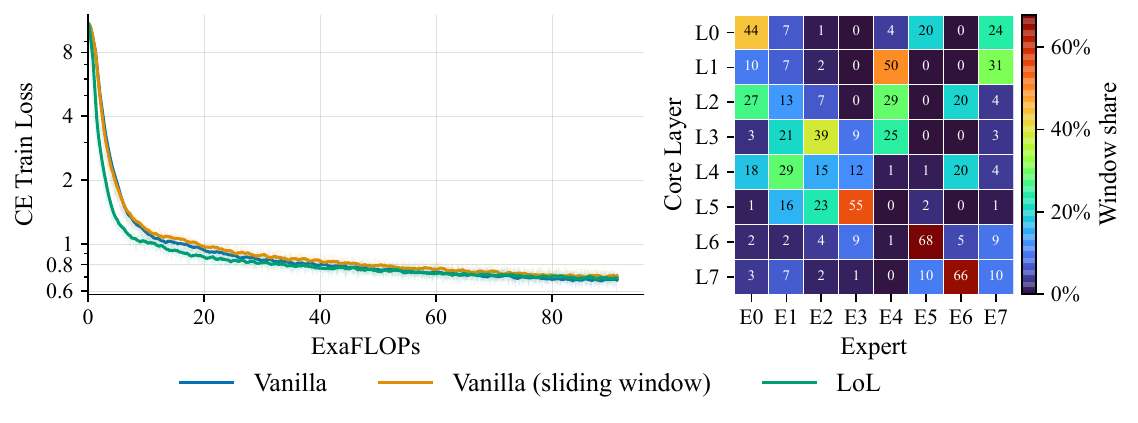}
    \caption{\textbf{\fullname is competitive with vanilla dense models.}
    Left: Cross-entropy training loss over training FLOPs for \(1B\) models. We see growing gives \name an early loss advantage and all models finish with comparable training loss.
    Right: Heatmap of expert utilization of one \name model. We see experts mainly tend to be used in groups of neighboring depths, but some are used throughout the model.
    }
    \label{fig:math}
\end{figure}

\subsubsection{Mazes}
Finally, we study mazes, where all five strategies share the same trace syntax, a list of visited cells, and only differ in the order in which they visit cells.
We solve loop-free mazes on an \(8\times 8\) grid using: BFS \citep{cormen1990introduction}, DFS \citep{cormen1990introduction}, A* \citep{A-Star}, bidirectional search \citep{pohl-bidir}, and dead-end filling.
See Appendix \Cref{app-tab:maze-data} for examples of each strategy.

We then repeat the sudoku procedure for mazes, extracting each algorithm's mode path. 
Unlike the strategies used for sudoku, the maze solving strategies all share the same output syntax and differ in the order in which they visit cells.
We therefore ask whether \name can learn these different exploration orders by changing the order of its own experts.
\Cref{fig:router} (right) shows that forcing \name to follow a particular computational path produces the requested algorithm's trace with nearly \(100\%\) accuracy, even when all strategies share the same output syntax.
Appendix \Cref{app-fig:maze_results} shows that \name also solves unseen mazes with close to \(100\%\) accuracy when given the strategy description in context, without path forcing.

\subsection{Mathematical Reasoning at the 1B Scale}\label{subsec:exp-math}
Next, we scale training to \(1B\) parameters with models that are \(16\) layers deep.
For \name, we divide the \(16\) layers into \(4\) prelude layers, \(4\) coda layers, and a depth of \(8\) in the routed core.
We train on a mixture of Nemotron \citep{mahabadi2025nemotron} and OpenMath \citep{toshniwal2024openmath2} data.
% The objective is to study whether learned layer selection remains competitive with a conventional fixed order Transformer at a larger scale.
% We focus on vanilla dense baselines because they retain the same basic Transformer building blocks and provide a direct reference under matched parameter and training-compute budgets.
We also grow the \name models using a \(1-sqrt\) schedule for \(75\%\) of training, following \citet{mcleish2025teaching}.
We train for approximately \(90\) ExaFLOPs; this equates to approximately \(20B\) tokens for the most efficient model.

\begin{table}[t!]
\centering
\small
\begin{tabular}{lccccc}
\toprule
Model & ARC-E & ARC-C & HellaSwag & PIQA & GSM8K \\
\midrule
Random & 25.0 & 25.0 & 25.0 & 50.0 & 0.0 \\
\midrule
Vanilla & $40.8 \pm 1.0$ & $28.2 \pm 1.3$ & $31.4 \pm 0.5$ & $56.6 \pm 1.2$ & $29.7 \pm 1.3$ \\
Vanilla (sliding window) & $49.5 \pm 1.0$ & $28.5 \pm 1.3$ & $31.3 \pm 0.5$ & $58.8 \pm 1.1$ & $27.5 \pm 1.2$ \\
\name & $50.3 \pm 1.0$ & $29.1 \pm 1.3$ & $31.2 \pm 0.5$ & $59.7 \pm 1.1$ & $27.9 \pm 1.2$ \\
\bottomrule
\end{tabular}
\caption{\textbf{\fullname is competitive across benchmarks.} We benchmark the same models shown in \Cref{fig:math} on ARC, HellaSwag, PIQA, and GSM8K. Performance is broadly comparable, except that the full-attention vanilla model scores substantially lower on ARC-Easy.}
\label{tab:math}
\end{table}

In \Cref{fig:math} (left), we see our models are competitive in terms of training loss with the parameter equivalent dense baselines.
We use both full attention and sliding window attention as baselines, where we set the size of the sliding window to be equal to the window size in the strided attention in \name.
In Appendix \Cref{app-fig:math} and \Cref{app-tab:math}, we include a comparison to the vanilla model with strided attention.
This model performs slightly worse than the vanilla and vanilla sliding window attention models.
We note that the math training data is synthetic; therefore, loss values in \Cref{fig:math} are lower than the FineWeb ones in other plots.
Moreover, in \Cref{fig:math} (right), we see our models are learning to use different experts at different routed depths.
In this setting, where the library size equals the routed core depth, we see clusters form where experts are often used across neighboring depths.
However, some experts are used across a wider range of depths.

In \Cref{tab:math}, we benchmark the same models on ARC-Easy \citep{clark2018think}, ARC-Challenge \citep{clark2018think}, HellaSwag \citep{zellers2019hellaswag}, PIQA \citep{bisk2020piqa}, and GSM8K \citep{cobbe2021training}.
We use \texttt{lm-evaluation-harness} \citep{eval-harness} and report the standard errors for each benchmark in the table.
We use these benchmarks as they are known to provide a stronger, more reliable signal at smaller model sizes and token counts \citep{magnusson2025datadecide,mcleish2026gemstones}.
We see almost all evaluations are competitive and remain within 2 standard errors of each other on all benchmarks, showing \name is competitive across a range of modern architectures.
The only exception is the vanilla model, which scores poorly on ARC-Easy.

\subsection{How Big Should Your Library Be?}\label{subsec:exp-sparse}
Library size $(L)$ determines how many distinct blocks are available, while core depth $(D)$ determines how many blocks are executed.
A small library requires reuse; a large library allows blocks to remain unused along a given routing path.
We compare the small-library regime with looped transformers and the large-library regime with sparse MoE models.

We use \(360M\) models for all comparisons, sweeping at least six learning rates, testing multiple learning rate schedules, and tuning the learning rates for Muon and AdamW separately.

\begin{figure}[t!]
    \centering
    \includegraphics[width=\linewidth]{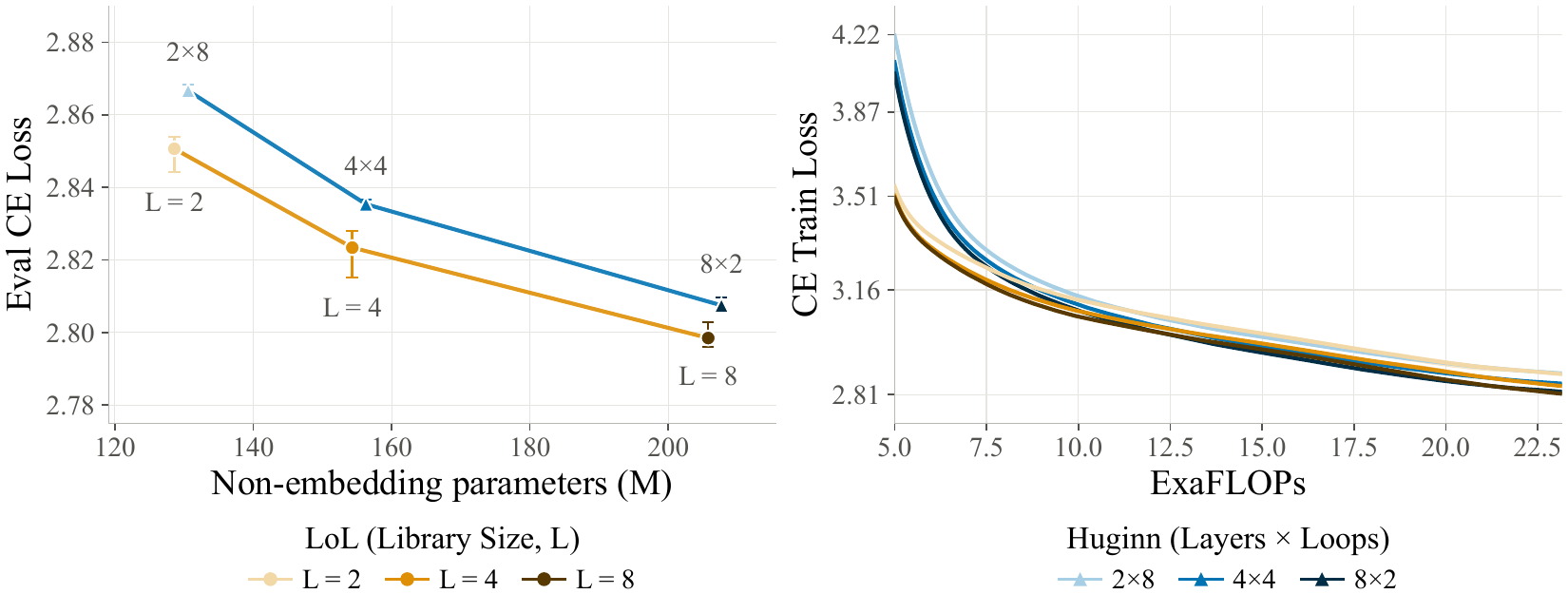}
    \caption{\textbf{\fullname is competitive with looped transformer models.}
    Left: Cross-entropy evaluation loss, matching the number of unique layers in the core block of the \name and looped transformer.
    We see \name performs competitively with looped transformer models and offers more flexibility in layer ordering.
    Error bars are min/max loss over \(3\) random seeds.
    We note larger variance for the \name models as changing the random seed changes the paths explored during training.
    Right: Cross-entropy training loss over training FLOPs, averaged over \(3\) random seeds per model.
    }
    \label{fig:huginn}
\end{figure}

\subsubsection{Comparison to Looped Transformers}
First, we compare to looped transformers.
% These models repeat layers multiple times in the forward pass, meaning that there is a high computational cost per unique parameter.
Following the analysis of \citet{huang2026towards}, we use the Huginn \citep{geiping2026scaling} architecture for our looped model baseline.
% We then compare models at multiple different unique parameter counts and reuse layers to achieve the same effective parameter count as our \(360M\) models.
We compare models at several distinct parameter counts, reusing layers to reach a total effective parameter count of \(360M\).
To ensure a fair comparison, we fix the library size in our \name models to match the size of the recurrent block in the Huginn models.
We train for \(10B\) tokens of FineWeb data \citep{penedo2024fineweb}.

In \Cref{fig:huginn}, we see that our \name models are competitive in terms of loss with the parameter matched looped models.
We note that the \name models have slightly higher variance in final loss over three seeds.
We hypothesize this is due to sensitivity to the initialization of the router causing different paths to be explored during training.
In Appendix \Cref{app-fig:looped_seeds}, we show the routing diversity for each \name model shown in \Cref{fig:huginn}.
Unlike Huginn, \name can offer sparse behavior by simply scaling the size of its library.

\subsubsection{Comparison to Sparse Mixture of Experts}
Finally, we compare to sparse mixture of experts models.
% We aim to use a sparsity ratio of \(3:2\), and achieve this ratio for the MoE models by using 3 experts per layer with 2 active.
% We match active parameters per token as closely as possible by increasing the expansion factor from hidden size to intermediate size in each individual expert of the MoE from that used in our regular \(360M\) model shape.
% For our \name models, we use a library size of \(24\) and depth of \(16\) to create the \(3:2\) sparsity equivalence.
% To match computation as closely as possible, in the \name models we use MoE layers for the prelude and coda, and we use strided attention in the MoE in the same layers in which the \name models use strided attention.
% We match active parameters per token as closely as possible.
We compare at a \(3:2\) sparsity ratio: the MoE model uses three experts per layer with two active, while \name uses 24 experts over 16 routed depths.
We increase the MoE experts’ intermediate expansion factor relative to our standard 360M model to match active parameters per token as closely as possible.
To align computation, \name uses MoE prelude and coda layers, and the MoE baseline uses strided attention at the same depths as \name.
We again use growing for the \name models and train all models for approximately \(17\) ExaFLOPs; this equates to approximately \(10B\) tokens for the most efficient model.

\begin{figure}[t!]
    \centering
    \includegraphics[width=\linewidth]{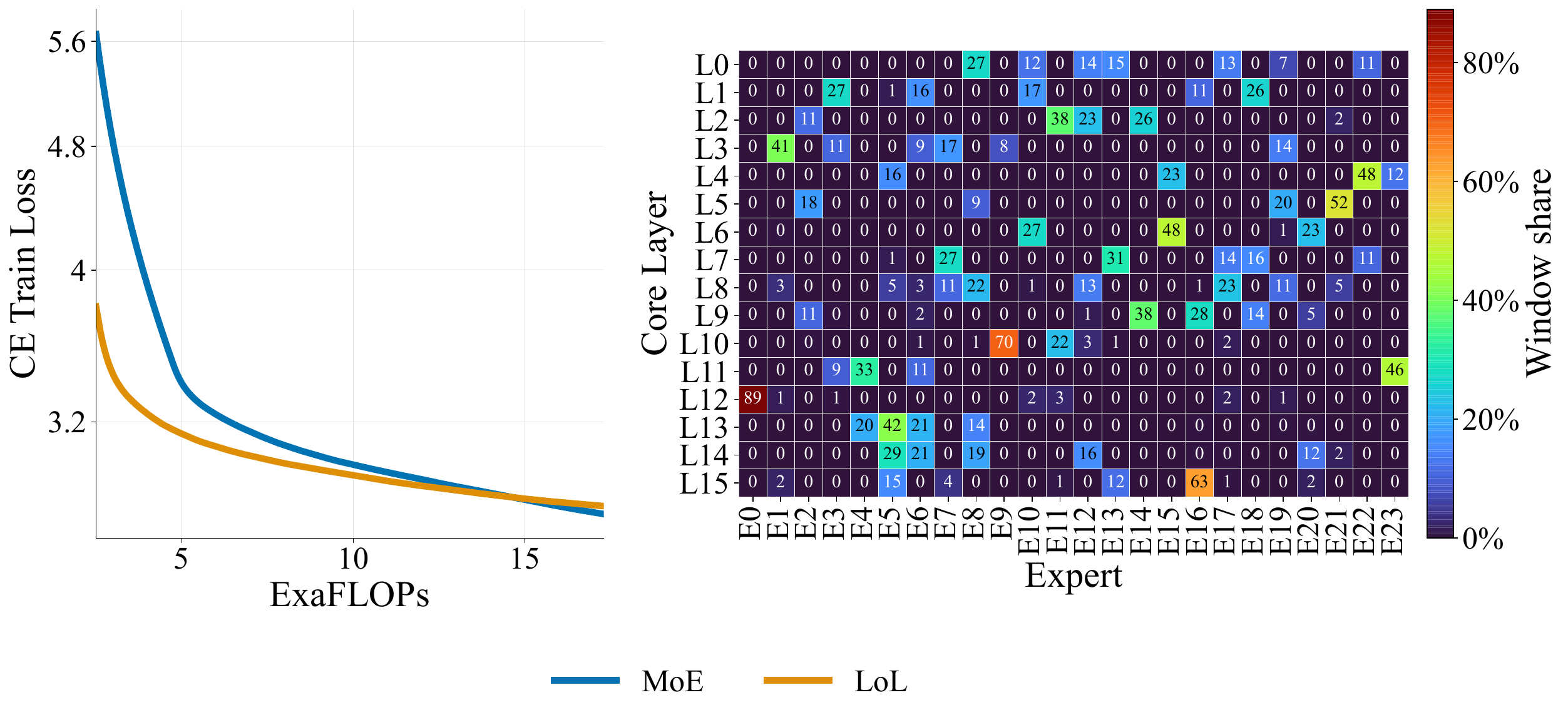}
    \caption{\textbf{\fullname is competitive with mixture of experts models.}
    Left: Cross-entropy training loss over training FLOPs, averaged over \(3\) random seeds. We see growing gives \name an early loss advantage and both models finish training with similar training losses.
    Right: Heatmap of expert utilization of one \name model. We see some experts are heavily used by specific routed depths or neighboring groups of depths, while others are used diversely over depths, showing \name learns to use the sparsity effectively.
    }
    \label{fig:moe}
\end{figure}

In \Cref{fig:moe} (left), the MoE baseline achieves a slightly lower training loss than \name, with a gap of approximately \(0.05\) (\(2.75\) vs. \(2.8\)).
We note it is common for higher cross-entropy loss to coexist with strong downstream performance in latent reasoning models \citep{saunshi2025reasoning,huang2026towards}.
In \Cref{fig:moe} (right), we visualize the expert utilization across routed depths on an evaluation set. 
Experts vary in depth specialization: some favor earlier or later depths, while others are used across depths, highlighting the flexibility of \name models on real-world language modeling data.

\section{Discussion}
\name generalizes the parameter reuse and sparsity of looped models, standard feedforward transformers, and MoEs into a single model family, allowing us to switch between these regimes simply by varying the library size.
While our design addresses several challenges we discuss above, this paper is an initial exploration into dynamic layer ordering, and we believe that future work will vastly improve on our design.
Moreover, the goal of our experiments is to show not merely that an architecture with dynamic layer ordering can achieve competitive performance, but also that it provides a new tool for model interpretability and control over a model’s reasoning strategies.  

We now enumerate several directions for future work that we believe are especially promising.
First, \name is built with controllability at the core.
We see in \Cref{fig:router} that we can find distinct routing paths underlying languages and algorithms, and we can intervene on routing paths at inference to control the model.
In \Cref{subsubsec:safety}, we show that routing features distinguish safe from unsafe text. 
Whether such features can also support interventions in chat settings, including behaviors related to lying, scheming, and safety \citep{deception2024_Park}, remains an open question.
Next, \name can in principle employ a larger computational depth at inference time than at training time, offering a mechanism for test-time scaling and increasing its capacity for latent reasoning. 
Finally, another promising direction is to combine test-time learning \citep{chen2026perk,hotsko2026code2lora,zhao2026fast} with our top-1 routing.  For example, instead of routing over full transformer blocks, one may route over a growing library of test-time learnable adapters for each layer.  This approach would allow the parametric memory to be available to all layers of the model.
% Finally, we believe heterogeneous libraries where different experts have different numbers of parameters or different architectural designs present an interesting future direction as they will allow the model to adaptively spend compute depending on the complexity of the task in a window.

%\textbf{Limitations.}
%The biggest limitation of our work is model size.
%Constrained by GPU access, we train models with up to \(1B\) parameters with token budgets at or above compute-optimal \citep{hoffmann2022training}.
%Future work may implement these models at larger sizes to show that benefits scale.
%Moreover, top-1 routers are prone to collapse during training.
%Although most of our runs avoid dead experts, we do see this in some cases and encourage future work on efficiently optimizing top-1 routers.

\section*{Acknowledgments}
We thank John Kirchenbauer for helpful feedback on earlier drafts of this paper.

\newpage

\bibliography{refs}
\bibliographystyle{iclr2027_conference}

\newpage
\section*{Table of Contents}
\DoToC 
\clearpage
\appendix

\section{Additional Architecture and Training Details}\label{app-sec:training-details}
\begin{algorithm}[t!]
\caption{\fullname{} forward pass}
\label{alg:forward-pass}
\begin{algorithmic}[1]
\setlength{\itemsep}{3pt}
\Require Input $\mathbf{x}\in V^n$, core depth $D$, window size $W$, stride $S$;
model parameters $\Theta=\{P,\{E_m\}_{m=1}^{L},\operatorname{Router},\{\mathbf{e}^{(\ell)}\}_{\ell=1}^{D},C\}$
\State $\mathbf{s}^{(0)} \gets P(\mathbf{x})$ \Comment{Full causal attention}
\For{$\ell=1,\ldots,D$}
    \Statex \textit{// Window preparation}
    \State $\delta_\ell \gets (\ell-1)S \bmod W$ % \Comment{Shift for the window boundaries}
    \State $I_k \gets \{\, i \in \{0,\ldots,n-1\} : \lfloor (i+\delta_\ell)/W \rfloor = k \,\}$, \; for $k=0,\ldots,K_\ell-1$
    \State $\boldsymbol{\mu}_k \gets \operatorname{Mean}\bigl(\mathbf{s}^{(\ell-1)}_{I_k}\bigr)$, \; for $k=0,\ldots,K_\ell-1$
    \Statex \textit{// Forward pass per window ($k$)}
    \For{$k=0,\ldots,K_\ell-1$}
        % \State $I_k \gets \{\, i \in \{0,\ldots,n-1\} : \lfloor (i+\delta_\ell)/W \rfloor = k \,\}$ \Comment{Current window}
        \If{$k=0$}
            \State $\mathbf{r}_k \gets \tfrac{1}{2}\bigl(\mathbf{e}^{(\ell)}+\mathbf{s}^{(\ell-1)}_0\bigr)$ \Comment{$\mathbf{e}^{(\ell)}$: learned vector for depth $\ell$}
        \Else
            \State $\mathbf{r}_k \gets \tfrac{1}{k}\sum_{j=0}^{k-1}\boldsymbol{\mu}_j$ \Comment{Mean of prior windows only}
        \EndIf
        \State $(\rho_k,\alpha_k) \gets \operatorname{Top1}\bigl(\operatorname{Router}(\mathbf{r}_k)\bigr)$ \Comment{$\rho_k$: selected expert; $\alpha_k$: expert weight}
        \State $\mathbf{y}_k \gets E_{\rho_k}\bigl(\mathbf{s}^{(\ell-1)}_{I_k}\bigr)$ \Comment{Causal attention within window}
        \State $\mathbf{s}^{(\ell)}_{I_k} \gets \mathbf{s}^{(\ell-1)}_{I_k} + \alpha_k\bigl(\mathbf{y}_k-\mathbf{s}^{(\ell-1)}_{I_k}\bigr)$
    \EndFor
\EndFor
\State $\mathbf{p} \gets \operatorname{softmax}\bigl(C(\mathbf{s}^{(D)})\bigr)$ \Comment{Full causal attention}
\State \Return $\mathbf{p}$
\end{algorithmic}
\end{algorithm}

Given a core depth of $D$, a library of $L$ experts, and an input sequence $\boldsymbol{x} \in V^{n}$, these three groups produce the output probabilities $\mathbf{p} \in \mathbb{R}^{n \times |V|}$ as shown in \Cref{alg:forward-pass}. 

Our main experiments are conducted at the 360M parameter scale based on the GPT-2 Medium architecture \citep{radford2019language}. 
Unlike GPT-2, our models use RoPE \citep{su2024roformer}, RMSNorm \citep{zhang2019root}, SwiGLU MLPs \citep{shazeer2020glu}, and bias free attention \citep{chowdhery2023palm}.
Our 1B model shape is based upon the Llama-3.2-1b shape \citep{meta_llama_3_2_announcement}, using the same architecture as the 360M model.
We use the GPT-NeoX tokenizer \citep{black2022gpt} for our predominantly English FineWeb training experiments and the larger OLMo3 \citep{olmo2025olmo} tokenizer for our multilingual experiments.
For our toy sudoku and maze experiments, we use small tokenizers specifically for the vocabulary of each task.
For natural language experiments, we use \(W=128\) and \(S=64\), with a total context length of \(1024\) tokens. For sudoku and mazes, we set the first window to span the question and strategy description. Subsequent windows use \(W=128\) and \(S=64\) for sudoku, and \(W=64\) and \(S=32\) for mazes.

To train the router, we use loss-free load balancing \citep{wang2024auxiliary}, adding a per-expert bias only for expert selection,
\(\rho_k = \arg\max_m \left(a_{k,m} + b_m\right),\)
where $a_{k,m}$ is the softmax probability of expert $m$ before adding the balancing bias. We weight the selected expert using this probability,
\(\alpha_k = a_{k,\rho_k}.\)
As in MoE routing \citep{shazeer2017outrageously,fedus2022switch}, this retains a differentiable path from the loss to the router through $\alpha_k$, without differentiating through the discrete selection.

For our \name models, we apply a learned, depth-dependent scale and bias after each normalization in the transformer block. 
Specifically, in place of the regular normalization in the transformer block, we use:
\(
\boldsymbol{\gamma}_{s} \odot \operatorname{Norm}_{s}(\mathbf{x}) + \boldsymbol{\beta}_{s},
\qquad s \in \{\mathrm{attn}, \mathrm{ffn}\},
\)
where \(\boldsymbol{\gamma}\) and \(\boldsymbol{\beta}\) are depth-dependent.

We use the Muon optimizer \citep{jordan2024muon} with AdamW \citep{kingma2014adam,loshchilov2017decoupled}.
We note we use PyTorch Muon with \texttt{adjust\_lr\_fn=match\_rms\_adamw}, and all learning rates should be taken in this context.
We use loss-free load balancing \citep{wang2024auxiliary} to train routers.
We use document blocked attention, meaning that tokens in one document never attend to tokens in another document in any experiment.

\section{Additional Architecture Ablations}\label{app-sec:arch-abl}
\subsection{\fullname}\label{app-sec:arch-abl-lol}
\begin{figure}[ht!]
    \centering
     \includegraphics[width=1\linewidth]{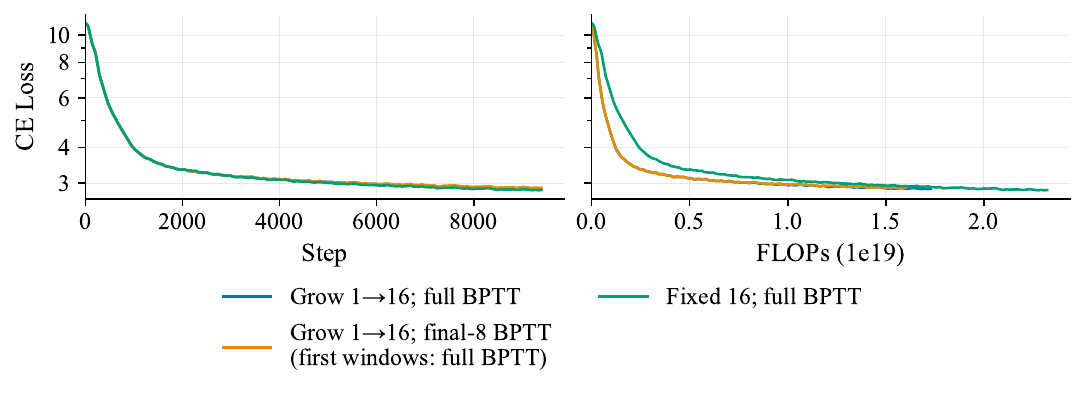}
    \caption{Increasing depth over training saves FLOPs with only a minor increase in final loss.}
    \label{app-fig:arch-abl-lol-backprop}
\end{figure}

\begin{figure}[ht!]
    \centering
    \includegraphics[width=1\linewidth]{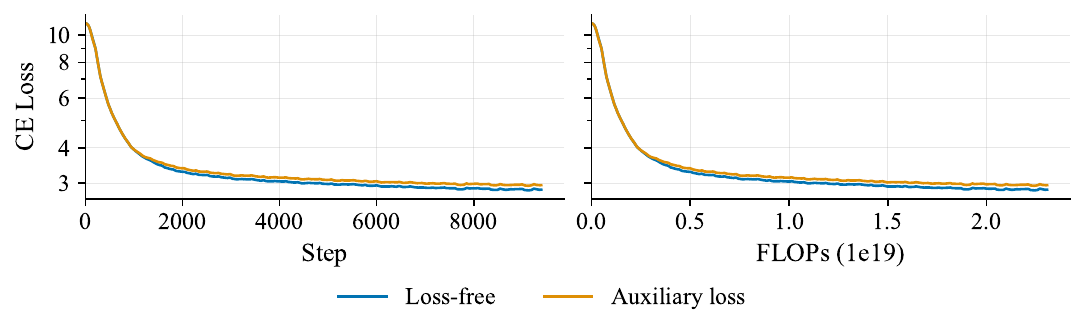}
    \caption{Loss-free load balancing \citep{wang2024auxiliary} achieves a lower final cross-entropy loss when compared to auxiliary loss load balancing methods.}
    \label{app-fig:arch-abl-lol-load}
\end{figure}

\begin{figure}[ht!]
    \centering
    \includegraphics[width=1\linewidth]{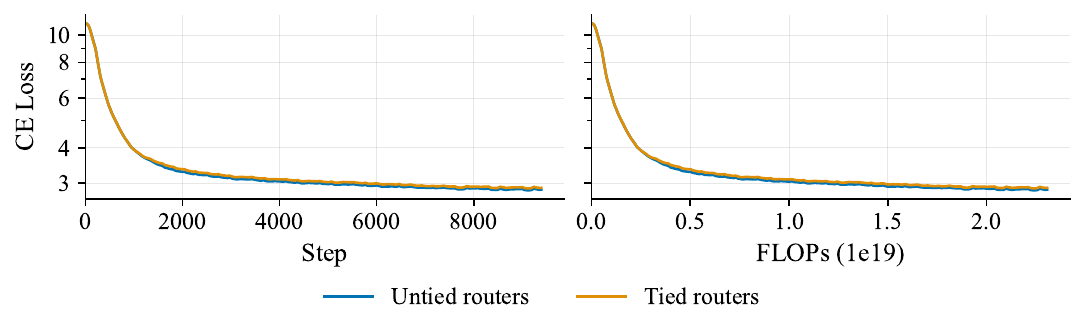}
    \caption{Sharing router weights across routed depths causes minimal change in cross-entropy loss.}
    \label{app-fig:arch-abl-lol-router-tie}
\end{figure}

In \Cref{app-fig:arch-abl-lol-backprop}, we ablate the impact of growing \name during training.
In \Cref{app-fig:arch-abl-lol-load}, we ablate the impact of auxiliary loss versus loss-free load balancing.
In \Cref{app-fig:arch-abl-lol-router-tie}, we see sharing router weights across routed depths has minimal impact on final training loss.

\subsection{Baselines}\label{app-sec:arch-abl-vanilla}
\begin{figure}[ht!]
    \centering
    \includegraphics[width=1\linewidth]{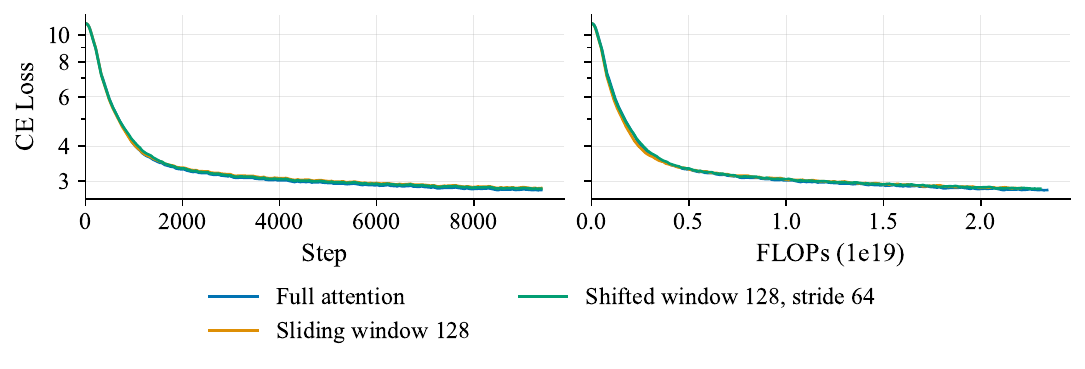}
    \caption{Varying the attention pattern for dense fixed-depth models, we see full attention achieves the lowest loss.}
    \label{app-fig:arch-abl-vanilla-attn}
\end{figure}

In \Cref{app-fig:arch-abl-vanilla-attn}, we see varying attention patterns in vanilla models has a small impact on training loss.

\section{Additional Experimental Information}
\subsection{Multilingual}
\begin{figure}[H]
    \centering
    \includegraphics[width=1\linewidth]{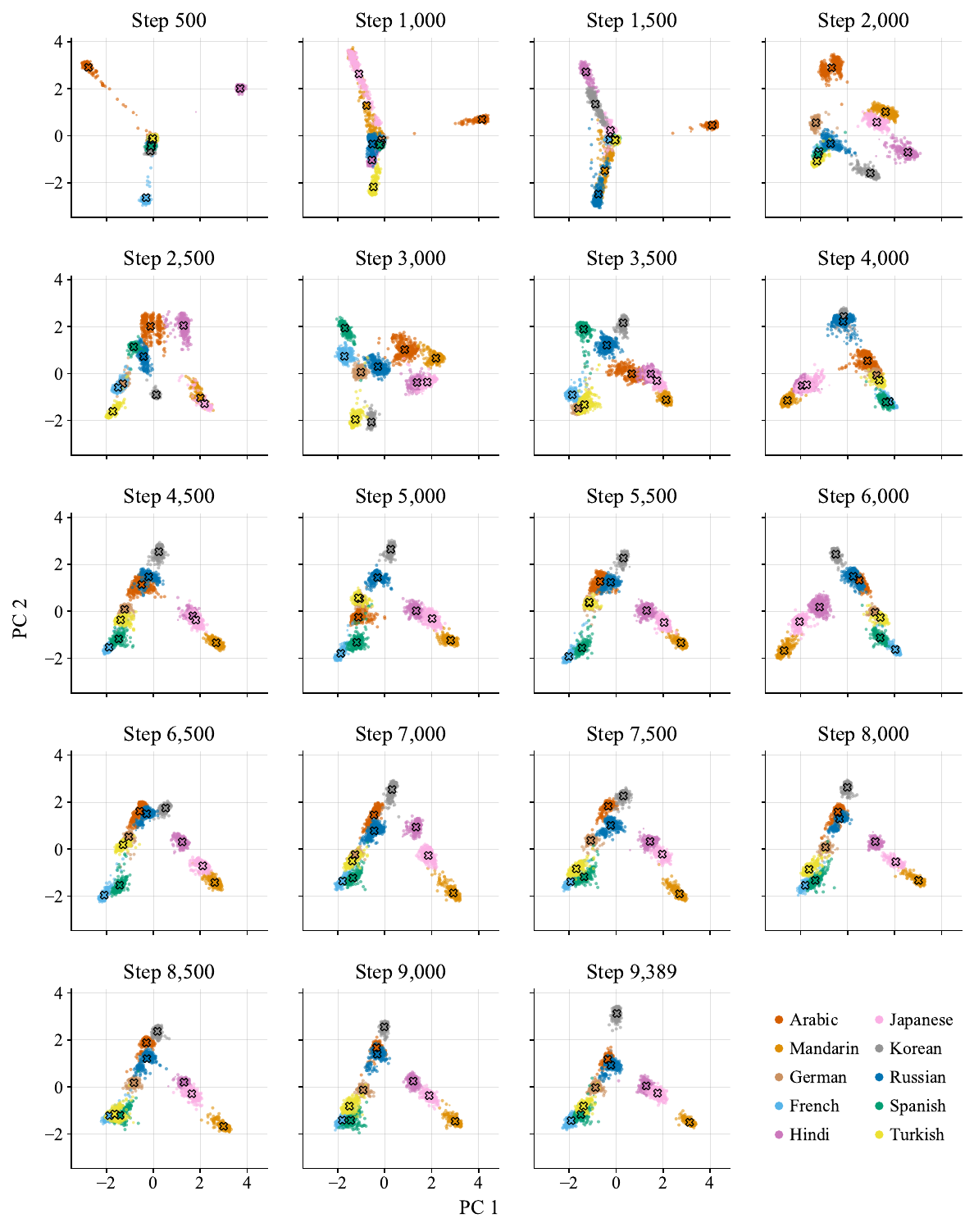}
    \caption{Evolution of multilingual routing during training.}
    \label{app-fig:multilang}
\end{figure}

\subsection{Safety}\label{app-sec:safety}
To confirm the safety classifier in \Cref{subsubsec:safety} is using predictive information in routing, we evaluate several readouts and a label-shuffling control.
We report the mean AUROC over five cross-validation folds of the development set.
In \Cref{app-tab:safety-readouts} we analyze: 
\begin{itemize}
    \item Ridge regression on the single top-1 expert probability at every depth.
    \item Single winning-expert probability where we directly rank windows by the negative winning-expert probability at a depth selected on the training folds.
    \item Expert-frequency log-likelihood ratio where we estimate class-conditional expert frequencies from training data.
\end{itemize}
The simpler readouts recover much of the top-1 ridge probe's performance, while shuffling labels at the document level reduces ridge performance to chance.
We note that this data is labeled by a classifier trained by \citet{maini2026safety}; hence, our results measure agreement with those noisy labels.

\begin{table}[ht]
\centering
\small
\begin{tabular}{lc}
\toprule
Readout or control & AUROC \\
\midrule
Top-1 ridge & 0.8754 \\
Single winning-expert probability & 0.8244 \\
Expert-frequency log-likelihood ratio & 0.8300 \\
Shuffled labels (99 permutations, mean / max) & 0.5014 / 0.5231 \\
\bottomrule
\end{tabular}
\caption{Development-set cross-validation AUROCs for safety readouts and the document-level label-shuffling control.}
\label{app-tab:safety-readouts}
\end{table}

\subsection{Sudoku}
To construct our dataset, we generate sudoku grids by randomized depth-first filling under the rules of sudoku.
We then retain between \(30\) and \(36\) digits in the grid, ensuring it is still solvable by all strategies.
Each puzzle is solved using all five strategies, and every trace becomes one sequence of the dataset.
In total, we have \(20\)M sequences, which equates to approximately \(10B\) training tokens, after deduplication.

\begin{table}[H]
    \centering
    \resizebox{\linewidth}{!}{
    \begin{tabular}{ll}
        \toprule
        Strategy & Solving Step Format\\
        \midrule
        Repeated row-major naked-single sweeps. & \texttt{r5c5 = \{ 5 \}}\\
        Hidden singles, scanning rows, columns, then boxes. & \texttt{row3 5 -> r3c7} \\
        Digit-major crosshatching over boxes. & \texttt{3 box4 -> r6c3}\\
        First-in-first-out propagation of forced cells, recording newly forced peers. & \texttt{set r5c5 = 5 : r5c2 -> \{ 2 \}}\\
        Deterministic DFS backtracking: first empty cell, digits ascending. & \texttt{r1c3 = 1 x}\\
        \bottomrule
    \end{tabular}
    }
    \caption{Sudoku data strategies and data formatting.}
    \label{app-tab:sudoku-data}
\end{table}

\begin{figure}[H]
    \centering
    \includegraphics[width=1\linewidth]{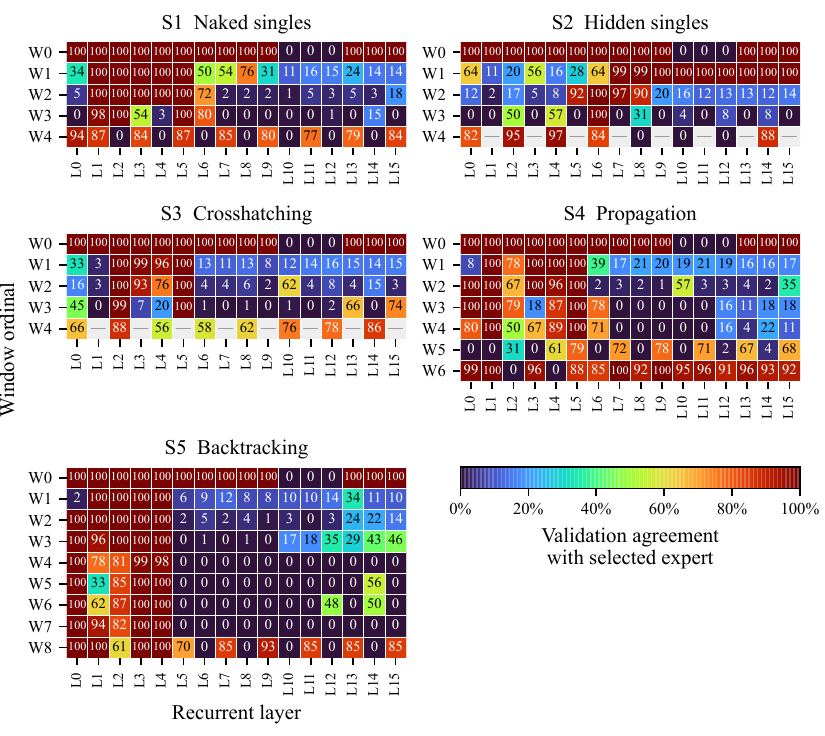}
    \caption{Routing agreement on sudoku. Each panel shows one strategy, with window positions as rows and routed depths \(k\) as columns. Each cell gives the percentage of observed routing decisions that agree with the mode-path expert at that position. Gray cells indicate positions with no observations.}
    \label{app-fig:sudoku_agree}
\end{figure}

\begin{figure}[H]
    \centering
    \includegraphics[width=1\linewidth]{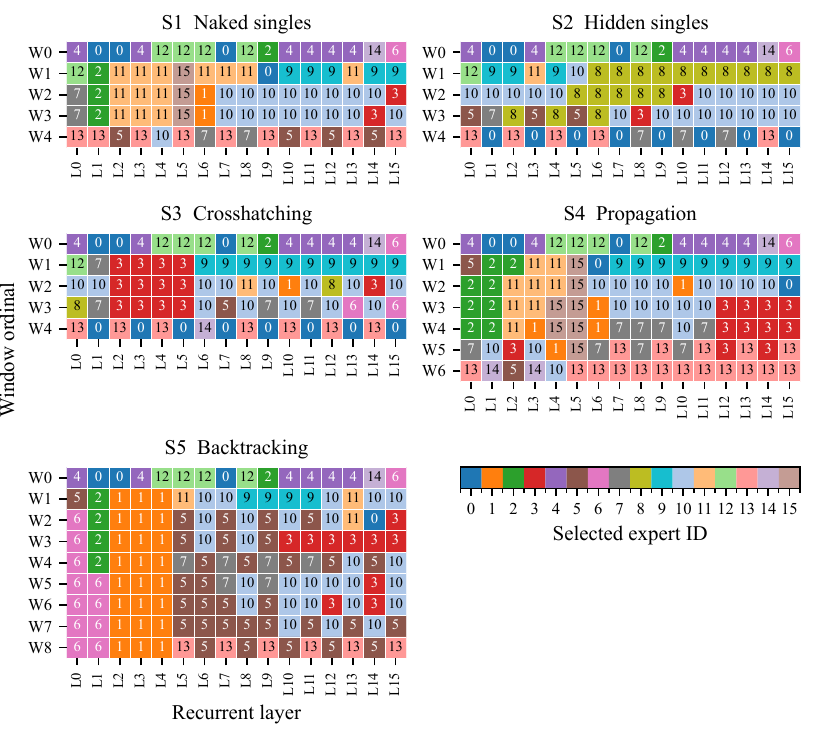}
    \caption{Mode routing paths on sudoku. Each panel shows one strategy, with window positions as rows and routed depths \(k\) as columns. Cell labels and colors identify the expert assigned to each (strategy, window position, routed depth) tuple.}
    \label{app-fig:sudoku_forced_path}
\end{figure}
\begin{figure}[H]
    \centering
    \includegraphics[width=0.8\linewidth]{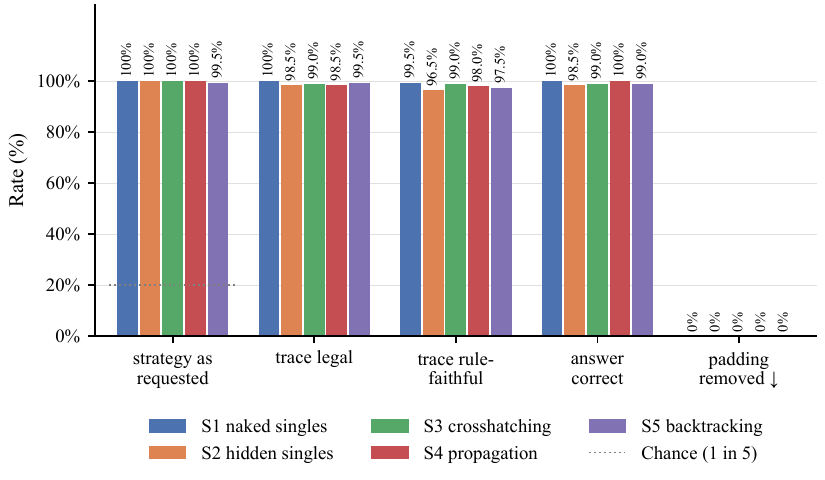}
    \caption{Strategy adherence and sudoku solution accuracy under forced routing.}
    \label{app-fig:sudoku_results}
\end{figure}
In \Cref{app-fig:sudoku_results}, we report different evaluation metrics:
\begin{itemize}
    \item \textbf{Strategy as requested:} the percentage of outputs whose solving steps all use the requested strategy's syntax.

    \item \textbf{Trace legal:} the percentage of outputs whose placements are all legal and lead to the true solution. Traces that do not match any strategy count as failures.

    \item \textbf{Trace rule-faithful:} the percentage of outputs that pass the legality check and correctly apply every claimed deduction.

    \item \textbf{Answer correct:} the percentage of outputs whose final 81 digit grid, after the answer marker, matches the true solution.

    \item \textbf{Padding removed \(\downarrow\):} the fraction of tokens before the end-of-sequence token that are padding, averaged across outputs.
\end{itemize}
We see that \name, with close to 100\% accuracy, uses the expected strategy with legal traces and succeeds in solving the sudoku grid without adding any unneeded padding. 

\subsubsection{Sudoku Data Example}

\begin{figure}[ht!]
    \centering
    \includegraphics[width=\linewidth]{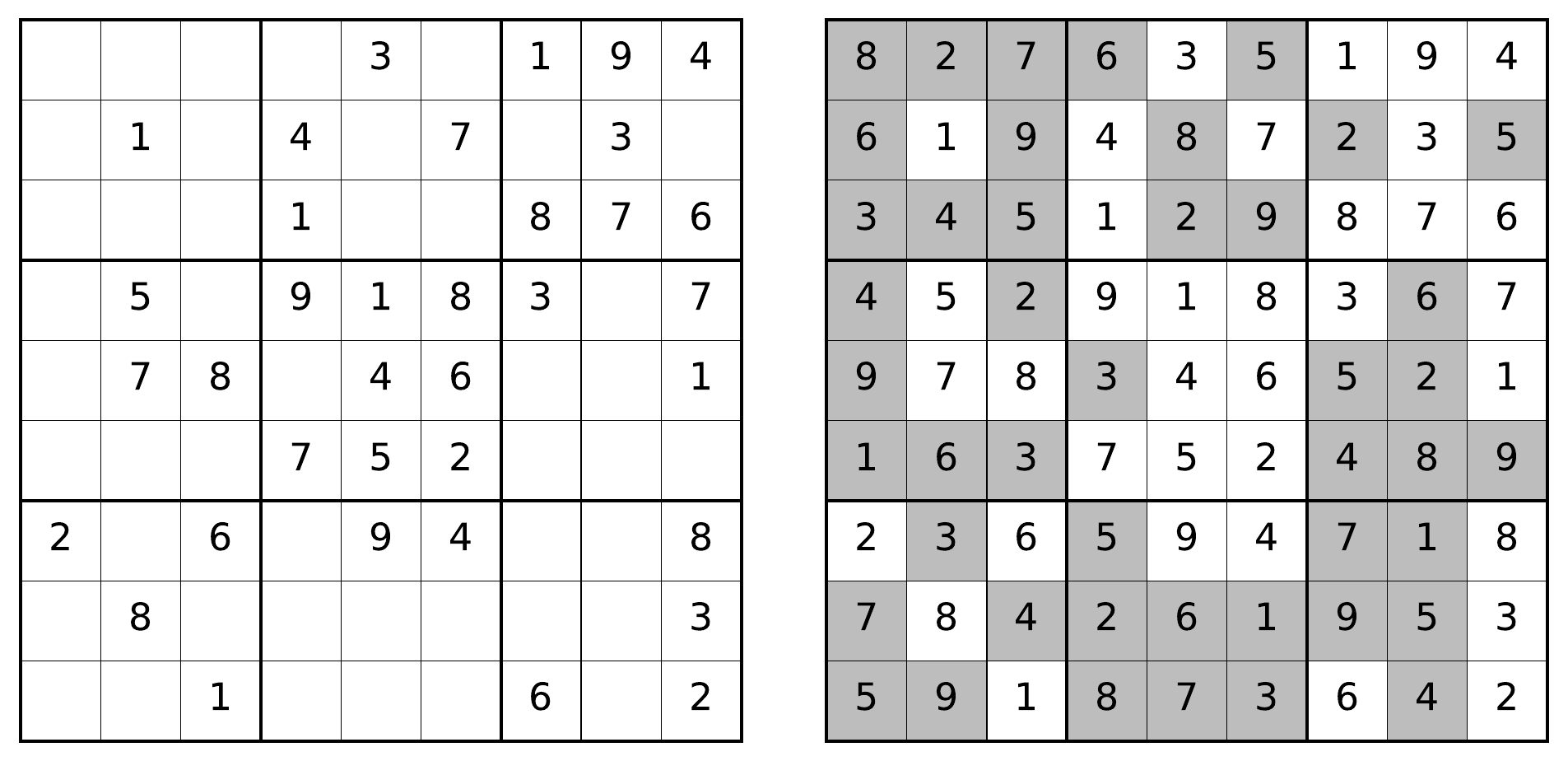}
    \caption{Example of sudoku data. We input the grid into the model as a sequence, as shown below.}
    \label{app-fig:sudoku_example}
\end{figure}
Our data is formatted as {\footnotesize\raggedright\noindent\texttt{Q : \textless{}Input Grid\textgreater{} [Sx] R : \textless{}Solving Steps\textgreater{} A : \textless{}Final Grid\textgreater{} \textless{}eos\textgreater{}}\par}.
In \Cref{app-fig:sudoku_example}, we visualize an example grid and below show the sequences to solve this grid using each strategy.

\par\medskip\noindent\textbf{S1: Naked singles}\par\smallskip
\noindent\fbox{\parbox{\dimexpr\linewidth-2\fboxsep-2\fboxrule\relax}{\footnotesize\raggedright\ttfamily Q : . . . . 3 . 1 9 4 . 1 . 4 . 7 . 3 . . . . 1 . . 8 7 6 . 5 . 9 1 8 3 . 7 . 7 8 . 4 6 . . 1 . . . 7 5 2 . . . 2 . 6 . 9 4 . . 8 . 8 . . . . . . 3 . . 1 . . . 6 . 2 [ S1 ] R : r1c6 = \{ 5 \} ; r2c9 = \{ 5 \} ; r3c5 = \{ 2 \} ; r3c6 = \{ 9 \} ; r5c4 = \{ 3 \} ; r6c9 = \{ 9 \} ; r7c2 = \{ 3 \} ; r7c4 = \{ 5 \} ; r7c7 = \{ 7 \} ; r7c8 = \{ 1 \} ; r8c6 = \{ 1 \} ; r9c4 = \{ 8 \} ; r9c5 = \{ 7 \} ; r9c6 = \{ 3 \} ; r1c4 = \{ 6 \} ; r2c5 = \{ 8 \} ; r2c7 = \{ 2 \} ; r3c2 = \{ 4 \} ; r5c1 = \{ 9 \} ; r5c7 = \{ 5 \} ; r5c8 = \{ 2 \} ; r6c2 = \{ 6 \} ; r6c7 = \{ 4 \} ; r6c8 = \{ 8 \} ; r8c4 = \{ 2 \} ; r8c5 = \{ 6 \} ; r8c7 = \{ 9 \} ; r9c2 = \{ 9 \} ; r1c2 = \{ 2 \} ; r1c3 = \{ 7 \} ; r2c1 = \{ 6 \} ; r2c3 = \{ 9 \} ; r4c1 = \{ 4 \} ; r4c3 = \{ 2 \} ; r4c8 = \{ 6 \} ; r6c3 = \{ 3 \} ; r9c1 = \{ 5 \} ; r9c8 = \{ 4 \} ; r1c1 = \{ 8 \} ; r3c1 = \{ 3 \} ; r3c3 = \{ 5 \} ; r6c1 = \{ 1 \} ; r8c1 = \{ 7 \} ; r8c3 = \{ 4 \} ; r8c8 = \{ 5 \} A : 8 2 7 6 3 5 1 9 4 6 1 9 4 8 7 2 3 5 3 4 5 1 2 9 8 7 6 4 5 2 9 1 8 3 6 7 9 7 8 3 4 6 5 2 1 1 6 3 7 5 2 4 8 9 2 3 6 5 9 4 7 1 8 7 8 4 2 6 1 9 5 3 5 9 1 8 7 3 6 4 2 \textless{}eos\textgreater{}\par}}

\par\medskip\noindent\textbf{S2: Hidden singles}\par\smallskip
\noindent\fbox{\parbox{\dimexpr\linewidth-2\fboxsep-2\fboxrule\relax}{\footnotesize\raggedright\ttfamily Q : . . . . 3 . 1 9 4 . 1 . 4 . 7 . 3 . . . . 1 . . 8 7 6 . 5 . 9 1 8 3 . 7 . 7 8 . 4 6 . . 1 . . . 7 5 2 . . . 2 . 6 . 9 4 . . 8 . 8 . . . . . . 3 . . 1 . . . 6 . 2 [ S2 ] R : row6 1 -\textgreater{} r6c1 ; row6 8 -\textgreater{} r6c8 ; row7 1 -\textgreater{} r7c8 ; row7 7 -\textgreater{} r7c7 ; row8 1 -\textgreater{} r8c6 ; col6 3 -\textgreater{} r9c6 ; col6 9 -\textgreater{} r3c6 ; col8 6 -\textgreater{} r4c8 ; col9 5 -\textgreater{} r2c9 ; col9 9 -\textgreater{} r6c9 ; box3 2 -\textgreater{} r2c7 ; box4 2 -\textgreater{} r4c3 ; box4 6 -\textgreater{} r6c2 ; box4 9 -\textgreater{} r5c1 ; box5 3 -\textgreater{} r5c4 ; box6 2 -\textgreater{} r5c8 ; box6 4 -\textgreater{} r6c7 ; box6 5 -\textgreater{} r5c7 ; box7 3 -\textgreater{} r7c2 ; box9 9 -\textgreater{} r8c7 ; row2 9 -\textgreater{} r2c3 ; row4 4 -\textgreater{} r4c1 ; row6 3 -\textgreater{} r6c3 ; row7 5 -\textgreater{} r7c4 ; row9 9 -\textgreater{} r9c2 ; col1 3 -\textgreater{} r3c1 ; col2 4 -\textgreater{} r3c2 ; col3 4 -\textgreater{} r8c3 ; col3 7 -\textgreater{} r1c3 ; col6 5 -\textgreater{} r1c6 ; col8 4 -\textgreater{} r9c8 ; col8 5 -\textgreater{} r8c8 ; box1 2 -\textgreater{} r1c2 ; box1 5 -\textgreater{} r3c3 ; box2 2 -\textgreater{} r3c5 ; box7 5 -\textgreater{} r9c1 ; box7 7 -\textgreater{} r8c1 ; box8 2 -\textgreater{} r8c4 ; box8 6 -\textgreater{} r8c5 ; box8 7 -\textgreater{} r9c5 ; box8 8 -\textgreater{} r9c4 ; row1 8 -\textgreater{} r1c1 ; row2 6 -\textgreater{} r2c1 ; row2 8 -\textgreater{} r2c5 ; col4 6 -\textgreater{} r1c4 A : 8 2 7 6 3 5 1 9 4 6 1 9 4 8 7 2 3 5 3 4 5 1 2 9 8 7 6 4 5 2 9 1 8 3 6 7 9 7 8 3 4 6 5 2 1 1 6 3 7 5 2 4 8 9 2 3 6 5 9 4 7 1 8 7 8 4 2 6 1 9 5 3 5 9 1 8 7 3 6 4 2 \textless{}eos\textgreater{}\par}}

\par\medskip\noindent\textbf{S3: Crosshatching}\par\smallskip
\noindent\fbox{\parbox{\dimexpr\linewidth-2\fboxsep-2\fboxrule\relax}{\footnotesize\raggedright\ttfamily Q : . . . . 3 . 1 9 4 . 1 . 4 . 7 . 3 . . . . 1 . . 8 7 6 . 5 . 9 1 8 3 . 7 . 7 8 . 4 6 . . 1 . . . 7 5 2 . . . 2 . 6 . 9 4 . . 8 . 8 . . . . . . 3 . . 1 . . . 6 . 2 [ S3 ] R : 1 box4 -\textgreater{} r6c1 ; 1 box8 -\textgreater{} r8c6 ; 1 box9 -\textgreater{} r7c8 ; 2 box3 -\textgreater{} r2c7 ; 2 box4 -\textgreater{} r4c3 ; 2 box6 -\textgreater{} r5c8 ; 3 box5 -\textgreater{} r5c4 ; 3 box8 -\textgreater{} r9c6 ; 5 box3 -\textgreater{} r2c9 ; 5 box6 -\textgreater{} r5c7 ; 8 box6 -\textgreater{} r6c8 ; 9 box2 -\textgreater{} r3c6 ; 9 box9 -\textgreater{} r8c7 ; 3 box7 -\textgreater{} r7c2 ; 6 box6 -\textgreater{} r4c8 ; 7 box9 -\textgreater{} r7c7 ; 9 box6 -\textgreater{} r6c9 ; 3 box4 -\textgreater{} r6c3 ; 4 box6 -\textgreater{} r6c7 ; 6 box4 -\textgreater{} r6c2 ; 9 box4 -\textgreater{} r5c1 ; 9 box7 -\textgreater{} r9c2 ; 3 box1 -\textgreater{} r3c1 ; 4 box4 -\textgreater{} r4c1 ; 4 box7 -\textgreater{} r8c3 ; 4 box9 -\textgreater{} r9c8 ; 5 box9 -\textgreater{} r8c8 ; 9 box1 -\textgreater{} r2c3 ; 4 box1 -\textgreater{} r3c2 ; 5 box7 -\textgreater{} r9c1 ; 5 box8 -\textgreater{} r7c4 ; 7 box7 -\textgreater{} r8c1 ; 7 box8 -\textgreater{} r9c5 ; 8 box8 -\textgreater{} r9c4 ; 2 box1 -\textgreater{} r1c2 ; 2 box2 -\textgreater{} r3c5 ; 2 box8 -\textgreater{} r8c4 ; 5 box2 -\textgreater{} r1c6 ; 6 box8 -\textgreater{} r8c5 ; 7 box1 -\textgreater{} r1c3 ; 8 box2 -\textgreater{} r2c5 ; 5 box1 -\textgreater{} r3c3 ; 6 box2 -\textgreater{} r1c4 ; 8 box1 -\textgreater{} r1c1 ; 6 box1 -\textgreater{} r2c1 A : 8 2 7 6 3 5 1 9 4 6 1 9 4 8 7 2 3 5 3 4 5 1 2 9 8 7 6 4 5 2 9 1 8 3 6 7 9 7 8 3 4 6 5 2 1 1 6 3 7 5 2 4 8 9 2 3 6 5 9 4 7 1 8 7 8 4 2 6 1 9 5 3 5 9 1 8 7 3 6 4 2 \textless{}eos\textgreater{}\par}}

\par\medskip\noindent\textbf{S4: Propagation}\par\smallskip
\noindent\fbox{\parbox{\dimexpr\linewidth-2\fboxsep-2\fboxrule\relax}{\footnotesize\raggedright\ttfamily Q : . . . . 3 . 1 9 4 . 1 . 4 . 7 . 3 . . . . 1 . . 8 7 6 . 5 . 9 1 8 3 . 7 . 7 8 . 4 6 . . 1 . . . 7 5 2 . . . 2 . 6 . 9 4 . . 8 . 8 . . . . . . 3 . . 1 . . . 6 . 2 [ S4 ] R : set r1c6 = 5 : r3c6 -\textgreater{} \{ 9 \} r8c6 -\textgreater{} \{ 1 \} r9c6 -\textgreater{} \{ 3 \} ; set r2c9 = 5 : r2c7 -\textgreater{} \{ 2 \} ; set r3c5 = 2 ; set r5c4 = 3 : r5c1 -\textgreater{} \{ 9 \} r7c4 -\textgreater{} \{ 5 \} ; set r6c9 = 9 : r6c7 -\textgreater{} \{ 4 \} ; set r7c2 = 3 ; set r3c6 = 9 : r3c2 -\textgreater{} \{ 4 \} ; set r8c6 = 1 ; set r9c6 = 3 ; set r2c7 = 2 : r2c3 -\textgreater{} \{ 9 \} r5c7 -\textgreater{} \{ 5 \} ; set r5c1 = 9 ; set r7c4 = 5 : r7c7 -\textgreater{} \{ 7 \} r7c8 -\textgreater{} \{ 1 \} r9c4 -\textgreater{} \{ 8 \} ; set r6c7 = 4 : r6c2 -\textgreater{} \{ 6 \} r6c3 -\textgreater{} \{ 3 \} ; set r3c2 = 4 : r9c2 -\textgreater{} \{ 9 \} ; set r2c3 = 9 ; set r5c7 = 5 : r5c8 -\textgreater{} \{ 2 \} ; set r7c7 = 7 : r8c7 -\textgreater{} \{ 9 \} ; set r7c8 = 1 ; set r9c4 = 8 : r1c4 -\textgreater{} \{ 6 \} r9c5 -\textgreater{} \{ 7 \} ; set r6c2 = 6 : r1c2 -\textgreater{} \{ 2 \} r4c1 -\textgreater{} \{ 4 \} r6c8 -\textgreater{} \{ 8 \} ; set r6c3 = 3 : r3c3 -\textgreater{} \{ 5 \} r6c1 -\textgreater{} \{ 1 \} ; set r9c2 = 9 ; set r5c8 = 2 : r4c8 -\textgreater{} \{ 6 \} ; set r8c7 = 9 ; set r1c4 = 6 : r2c5 -\textgreater{} \{ 8 \} r8c4 -\textgreater{} \{ 2 \} ; set r9c5 = 7 : r8c5 -\textgreater{} \{ 6 \} ; set r1c2 = 2 : r1c3 -\textgreater{} \{ 7 \} ; set r4c1 = 4 : r4c3 -\textgreater{} \{ 2 \} r9c1 -\textgreater{} \{ 5 \} ; set r6c8 = 8 ; set r3c3 = 5 : r3c1 -\textgreater{} \{ 3 \} ; set r6c1 = 1 ; set r4c8 = 6 ; set r2c5 = 8 : r2c1 -\textgreater{} \{ 6 \} ; set r8c4 = 2 ; set r8c5 = 6 ; set r1c3 = 7 : r1c1 -\textgreater{} \{ 8 \} r8c3 -\textgreater{} \{ 4 \} ; set r4c3 = 2 ; set r9c1 = 5 : r8c1 -\textgreater{} \{ 7 \} r9c8 -\textgreater{} \{ 4 \} ; set r3c1 = 3 ; set r2c1 = 6 ; set r1c1 = 8 ; set r8c3 = 4 : r8c8 -\textgreater{} \{ 5 \} ; set r8c1 = 7 ; set r9c8 = 4 ; set r8c8 = 5 A : 8 2 7 6 3 5 1 9 4 6 1 9 4 8 7 2 3 5 3 4 5 1 2 9 8 7 6 4 5 2 9 1 8 3 6 7 9 7 8 3 4 6 5 2 1 1 6 3 7 5 2 4 8 9 2 3 6 5 9 4 7 1 8 7 8 4 2 6 1 9 5 3 5 9 1 8 7 3 6 4 2 \textless{}eos\textgreater{}\par}}

\par\medskip\noindent\textbf{S5: Backtracking}\par\smallskip
\noindent\fbox{\parbox{\dimexpr\linewidth-2\fboxsep-2\fboxrule\relax}{\footnotesize\raggedright\ttfamily Q : . . . . 3 . 1 9 4 . 1 . 4 . 7 . 3 . . . . 1 . . 8 7 6 . 5 . 9 1 8 3 . 7 . 7 8 . 4 6 . . 1 . . . 7 5 2 . . . 2 . 6 . 9 4 . . 8 . 8 . . . . . . 3 . . 1 . . . 6 . 2 [ S5 ] R : r1c1 = 5 x ; r1c1 = 6 ; r1c2 = 2 ; r1c3 = 5 x ; r1c3 = 7 ; r1c4 = 5 x ; r1c4 = 8 ; r1c6 = 5 ; r2c1 = 5 x ; r2c1 = 8 ; r2c3 = 5 x ; r2c3 = 9 ; r2c5 = 2 x ; r2c5 = 6 ; r2c7 = 2 ; r2c9 = 5 ; r3c1 = 3 ; r3c2 = 4 ; r3c3 = 5 ; r3c5 = 2 ; r3c6 = 9 ; r4c1 = 4 ; r4c3 = 2 ; r4c8 = 6 ; r5c1 = 9 ; r5c4 = 3 ; r5c7 = 5 ; r5c8 = 2 ; r6c1 = 1 ; r6c2 = 3 x ; r6c2 = 6 ; r6c3 = 3 ; r6c7 = 4 ; r6c8 = 8 ; r6c9 = 9 ; r7c2 = 3 ; r7c4 = 5 x ; r7c4 back ; r7c2 back ; r6c9 back ; r6c8 back ; r6c7 = 9 x ; r6c7 back ; r6c3 back ; r6c2 back ; r6c1 back ; r5c8 back ; r5c7 back ; r5c4 back ; r5c1 back ; r4c8 back ; r4c3 back ; r4c1 back ; r3c6 back ; r3c5 back ; r3c3 back ; r3c2 back ; r3c1 = 4 x ; r3c1 = 5 ; r3c2 = 3 x ; r3c2 = 4 ; r3c3 = 3 ; r3c5 = 2 ; r3c6 = 9 ; r4c1 = 4 x ; r4c1 back ; r3c6 back ; r3c5 back ; r3c3 back ; r3c2 back ; r3c1 back ; r2c9 back ; r2c7 = 5 x ; r2c7 back ; r2c5 back ; r2c3 back ; r2c1 = 9 x ; r2c1 back ; r1c6 back ; r1c4 back ; r1c3 back ; r1c2 back ; r1c1 = 7 ; r1c2 = 2 ; r1c3 = 5 x ; r1c3 back ; r1c2 = 6 ; r1c3 = 2 x ; r1c3 = 5 x ; r1c3 back ; r1c2 back ; r1c1 = 8 ; r1c2 = 2 ; r1c3 = 5 x ; r1c3 = 7 ; r1c4 = 5 x ; r1c4 = 6 ; r1c6 = 5 ; r2c1 = 5 x ; r2c1 = 6 ; r2c3 = 5 x ; r2c3 = 9 ; r2c5 = 2 x ; r2c5 = 8 ; r2c7 = 2 ; r2c9 = 5 ; r3c1 = 3 ; r3c2 = 4 ; r3c3 = 5 ; r3c5 = 2 ; r3c6 = 9 ; r4c1 = 4 ; r4c3 = 2 ; r4c8 = 6 ; r5c1 = 9 ; r5c4 = 3 ; r5c7 = 5 ; r5c8 = 2 ; r6c1 = 1 ; r6c2 = 3 x ; r6c2 = 6 ; r6c3 = 3 ; r6c7 = 4 ; r6c8 = 8 ; r6c9 = 9 ; r7c2 = 3 ; r7c4 = 5 ; r7c7 = 7 ; r7c8 = 1 ; r8c1 = 5 ; r8c3 = 4 x ; r8c3 back ; r8c1 = 7 ; r8c3 = 4 ; r8c4 = 2 ; r8c5 = 6 ; r8c6 = 1 ; r8c7 = 9 ; r8c8 = 5 ; r9c1 = 5 ; r9c2 = 9 ; r9c4 = 8 ; r9c5 = 7 ; r9c6 = 3 ; r9c8 = 4 A : 8 2 7 6 3 5 1 9 4 6 1 9 4 8 7 2 3 5 3 4 5 1 2 9 8 7 6 4 5 2 9 1 8 3 6 7 9 7 8 3 4 6 5 2 1 1 6 3 7 5 2 4 8 9 2 3 6 5 9 4 7 1 8 7 8 4 2 6 1 9 5 3 5 9 1 8 7 3 6 4 2 \textless{}eos\textgreater{}\par}}

\subsection{Maze}
We generate loop-free mazes on an \(8\times 8\) grid using randomized recursive backtracking and randomly choose distinct start and goal cells.
We ensure all mazes have a unique solving trace per strategy and a solution path of at least \(12\) cells.
We also require DFS to backtrack at least once and BFS to expand at least three cells outside the solution path.
As in our sudoku experiments, we organize each example into a sequence of problem-solving steps, remove duplicate mazes, only allow the router to view the strategy description, and compute the loss only over the solving steps and final answer.
We train on approximately \(10B\) tokens of data made up of \(7.2\)M mazes and \(36\)M sequences, one per strategy for each maze.

\begin{table}[ht!]
    \centering
    \begin{tabular}{ll}
        \toprule
        Strategy & Example Solving Steps \\
        \midrule
        Breadth-first search in discovery order.
        & \texttt{r1c1 r1c2 r2c1 r1c3 r2c2 r3c1} \\

        Depth-first traversal with backtracking.
        & \texttt{r1c1 r1c2 r1c3 r1c4 r1c3 r1c2} \\

        A* search using Manhattan distance.
        & \texttt{r1c1 r1c2 r2c2 r2c3 r3c3 r3c4} \\

        Alternating BFS from the start and goal.
        & \texttt{r1c1 r8c8 r1c2 r8c7 r1c3 r8c6} \\

        Repeated row-major dead-end filling.
        & \texttt{r8c1 r8c2 r8c3 r1c8 r2c8 r3c8} \\
        \bottomrule
    \end{tabular}
    \caption{Maze data strategies and data formatting.}
    \label{app-tab:maze-data}
\end{table}

\begin{figure}[H]
    \centering
    \includegraphics[width=0.9\linewidth]{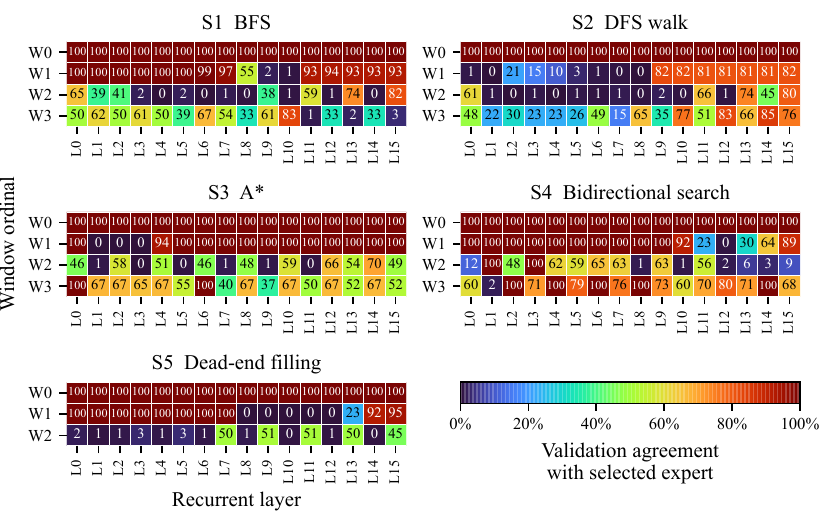}
  \caption{Routing agreement on mazes. Each panel shows one strategy, with window positions as rows and routed depths \(k\) as columns. Each cell gives the percentage of observed routing decisions that agree with the mode-path expert at that position. The gray row marks window W3 for dead-end filling; no validation observations are available for that window-strategy combination.}
    \label{app-fig:maze_agree}
\end{figure}

\begin{figure}[H]
    \centering
    \includegraphics[width=0.9\linewidth]{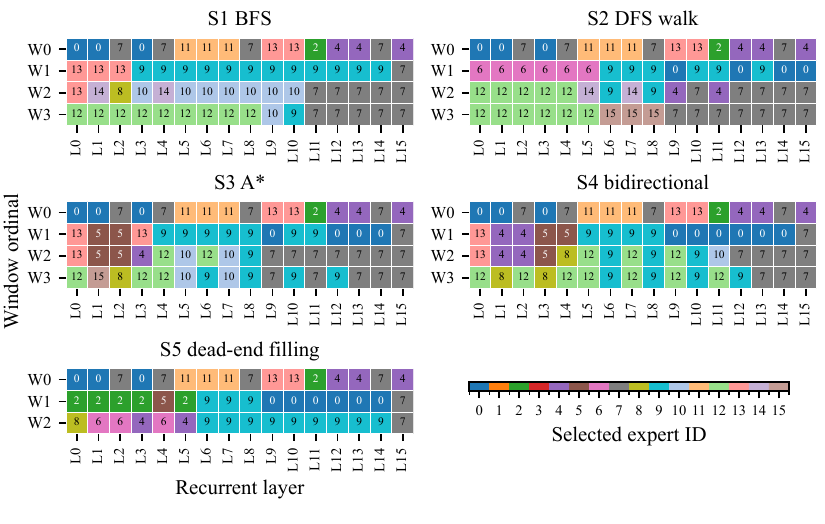}
    \caption{Mode routing paths on mazes. Each panel shows one strategy, with window positions as rows and routed depths \(k\) as columns. Cell labels and colors identify the expert with the highest mean router probability for each (strategy, window position, routed depth) tuple during validation. For dead-end filling, W3 has no validation observations and reuses the W2 expert sequence if reached.}
    \label{app-fig:maze_forced_path}
\end{figure}
\begin{figure}[H]
    \centering
    \includegraphics[width=1\linewidth]{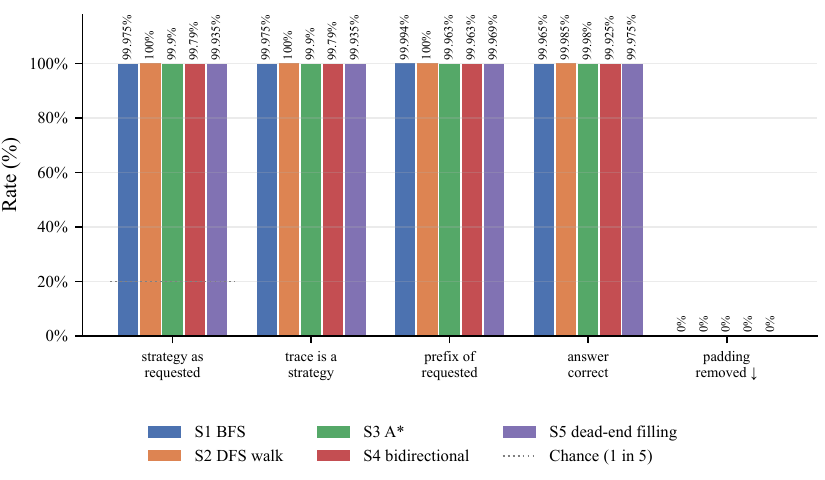}
    \caption{Strategy and maze solution generation under forced routing.}
    \label{app-fig:maze_results}
\end{figure}
The evaluations reported in the maze experiment are the same as those in the sudoku experiment.

\subsubsection{Maze Data Example}
\begin{figure}[ht!]
    \centering
    \includegraphics[width=\linewidth]{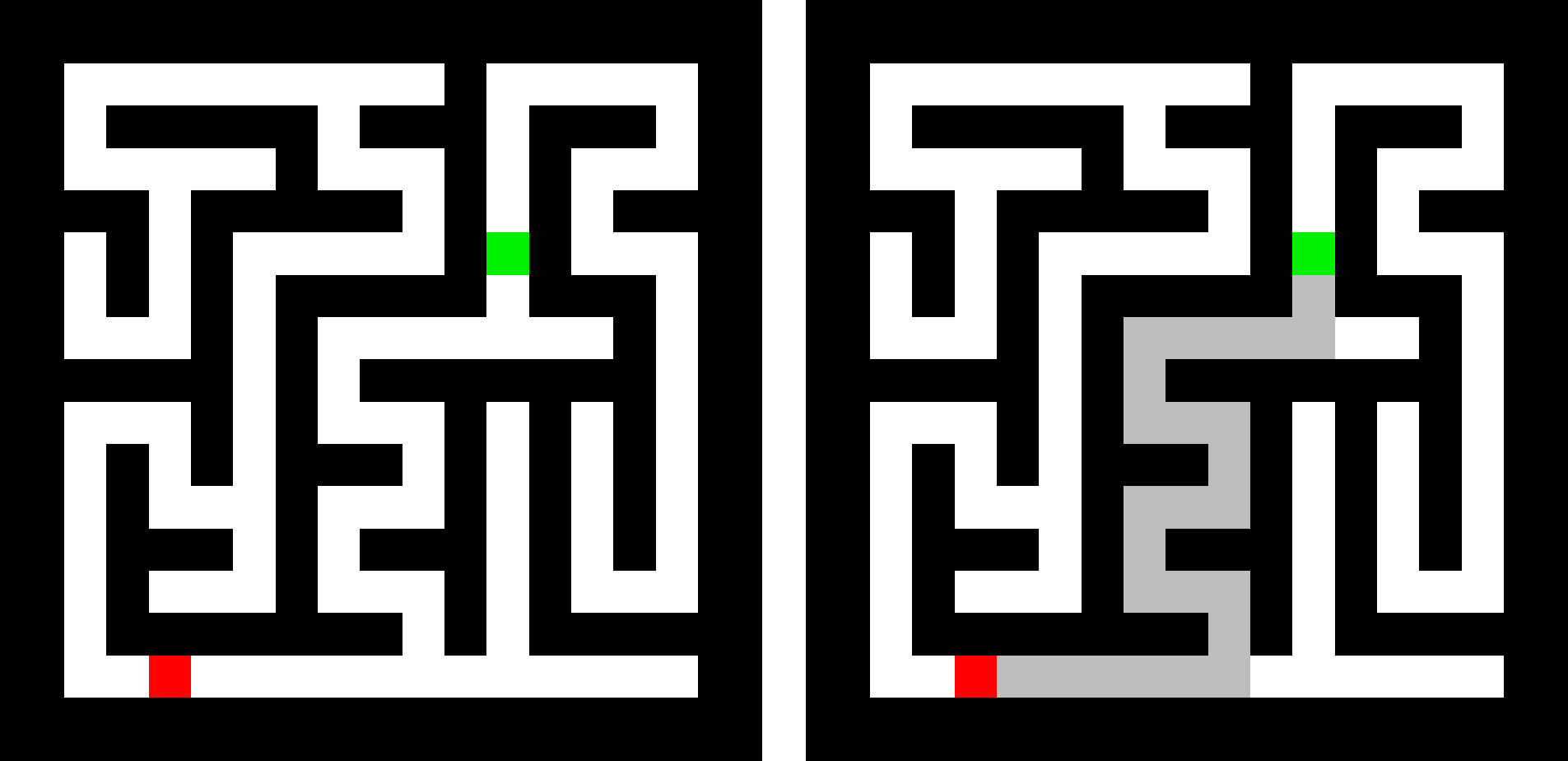}
    \caption{Example of maze data. We input the maze into the model as a sequence, as shown below.}
    \label{app-fig:maze_example}
\end{figure}
Our data is formatted as \texttt{Q : \textless{}Maze\textgreater{} start \textless{}Start Cell\textgreater{} goal \textless{}Goal Cell\textgreater{} [\textless{}Strategy Label\textgreater{}] R : \textless{}Search Trace\textgreater{} A : \textless{}Solution Path\textgreater{} \textless{}eos\textgreater{}}.
In \Cref{app-fig:maze_example}, we visualize an example maze and below show the sequences to solve this maze using each strategy.

\par\medskip\noindent\textbf{S1: BFS}\par\smallskip
\noindent\fbox{\parbox{\dimexpr\linewidth-2\fboxsep-2\fboxrule\relax}{\footnotesize\raggedright\ttfamily Q : r1c1 E S r1c2 E W r1c3 E W r1c4 E S W r1c5 W r1c6 E S r1c7 E W r1c8 S W r2c1 N E r2c2 E S W r2c3 W r2c4 N E r2c5 S W r2c6 N S r2c7 E S r2c8 N W r3c1 S r3c2 N S r3c3 E S r3c4 E W r3c5 N W r3c6 N S r3c7 N E r3c8 S W r4c1 N E r4c2 N W r4c3 N S r4c4 E S r4c5 E W r4c6 N E W r4c7 W r4c8 N S r5c1 E S r5c2 S W r5c3 N S r5c4 N E r5c5 S W r5c6 S r5c7 S r5c8 N S r6c1 N S r6c2 N E r6c3 N S W r6c4 E S r6c5 N W r6c6 N S r6c7 N S r6c8 N S r7c1 N S r7c2 E r7c3 N W r7c4 N E r7c5 S W r7c6 N S r7c7 N E r7c8 N W r8c1 N E r8c2 E W r8c3 E W r8c4 E W r8c5 N E W r8c6 N E W r8c7 E W r8c8 W start r3c6 goal r8c2 [ r1c1 r1c2 r2c1 r1c3 r2c2 r3c1 ] R : r3c6 r2c6 r4c6 r1c6 r4c7 r4c5 r1c7 r4c4 r1c8 r5c4 r2c8 r5c5 r2c7 r6c5 r3c7 r6c4 r3c8 r7c4 r4c8 r7c5 r5c8 r8c5 r6c8 r8c6 r8c4 r7c8 r7c6 r8c7 r8c3 r7c7 r6c6 r8c8 r8c2 A : r3c6 r4c6 r4c5 r4c4 r5c4 r5c5 r6c5 r6c4 r7c4 r7c5 r8c5 r8c4 r8c3 r8c2 \textless{}eos\textgreater{}\par}}

\par\medskip\noindent\textbf{S2: DFS walk}\par\smallskip
\noindent\fbox{\parbox{\dimexpr\linewidth-2\fboxsep-2\fboxrule\relax}{\footnotesize\raggedright\ttfamily Q : r1c1 E S r1c2 E W r1c3 E W r1c4 E S W r1c5 W r1c6 E S r1c7 E W r1c8 S W r2c1 N E r2c2 E S W r2c3 W r2c4 N E r2c5 S W r2c6 N S r2c7 E S r2c8 N W r3c1 S r3c2 N S r3c3 E S r3c4 E W r3c5 N W r3c6 N S r3c7 N E r3c8 S W r4c1 N E r4c2 N W r4c3 N S r4c4 E S r4c5 E W r4c6 N E W r4c7 W r4c8 N S r5c1 E S r5c2 S W r5c3 N S r5c4 N E r5c5 S W r5c6 S r5c7 S r5c8 N S r6c1 N S r6c2 N E r6c3 N S W r6c4 E S r6c5 N W r6c6 N S r6c7 N S r6c8 N S r7c1 N S r7c2 E r7c3 N W r7c4 N E r7c5 S W r7c6 N S r7c7 N E r7c8 N W r8c1 N E r8c2 E W r8c3 E W r8c4 E W r8c5 N E W r8c6 N E W r8c7 E W r8c8 W start r3c6 goal r8c2 [ r1c1 r1c2 r1c3 r1c4 r1c3 r1c2 ] R : r3c6 r2c6 r1c6 r1c7 r1c8 r2c8 r2c7 r3c7 r3c8 r4c8 r5c8 r6c8 r7c8 r7c7 r6c7 r5c7 r6c7 r7c7 r7c8 r6c8 r5c8 r4c8 r3c8 r3c7 r2c7 r2c8 r1c8 r1c7 r1c6 r2c6 r3c6 r4c6 r4c7 r4c6 r4c5 r4c4 r5c4 r5c5 r6c5 r6c4 r7c4 r7c5 r8c5 r8c6 r7c6 r6c6 r5c6 r6c6 r7c6 r8c6 r8c7 r8c8 r8c7 r8c6 r8c5 r8c4 r8c3 r8c2 A : r3c6 r4c6 r4c5 r4c4 r5c4 r5c5 r6c5 r6c4 r7c4 r7c5 r8c5 r8c4 r8c3 r8c2 \textless{}eos\textgreater{}\par}}

\par\medskip\noindent\textbf{S3: A*}\par\smallskip
\noindent\fbox{\parbox{\dimexpr\linewidth-2\fboxsep-2\fboxrule\relax}{\footnotesize\raggedright\ttfamily Q : r1c1 E S r1c2 E W r1c3 E W r1c4 E S W r1c5 W r1c6 E S r1c7 E W r1c8 S W r2c1 N E r2c2 E S W r2c3 W r2c4 N E r2c5 S W r2c6 N S r2c7 E S r2c8 N W r3c1 S r3c2 N S r3c3 E S r3c4 E W r3c5 N W r3c6 N S r3c7 N E r3c8 S W r4c1 N E r4c2 N W r4c3 N S r4c4 E S r4c5 E W r4c6 N E W r4c7 W r4c8 N S r5c1 E S r5c2 S W r5c3 N S r5c4 N E r5c5 S W r5c6 S r5c7 S r5c8 N S r6c1 N S r6c2 N E r6c3 N S W r6c4 E S r6c5 N W r6c6 N S r6c7 N S r6c8 N S r7c1 N S r7c2 E r7c3 N W r7c4 N E r7c5 S W r7c6 N S r7c7 N E r7c8 N W r8c1 N E r8c2 E W r8c3 E W r8c4 E W r8c5 N E W r8c6 N E W r8c7 E W r8c8 W start r3c6 goal r8c2 [ r1c1 r1c2 r2c2 r2c3 r3c3 r3c4 ] R : r3c6 r4c6 r4c5 r4c4 r5c4 r2c6 r4c7 r5c5 r6c5 r6c4 r7c4 r1c6 r7c5 r8c5 r8c4 r8c3 r8c2 A : r3c6 r4c6 r4c5 r4c4 r5c4 r5c5 r6c5 r6c4 r7c4 r7c5 r8c5 r8c4 r8c3 r8c2 \textless{}eos\textgreater{}\par}}

\par\medskip\noindent\textbf{S4: Bidirectional BFS}\par\smallskip
\noindent\fbox{\parbox{\dimexpr\linewidth-2\fboxsep-2\fboxrule\relax}{\footnotesize\raggedright\ttfamily Q : r1c1 E S r1c2 E W r1c3 E W r1c4 E S W r1c5 W r1c6 E S r1c7 E W r1c8 S W r2c1 N E r2c2 E S W r2c3 W r2c4 N E r2c5 S W r2c6 N S r2c7 E S r2c8 N W r3c1 S r3c2 N S r3c3 E S r3c4 E W r3c5 N W r3c6 N S r3c7 N E r3c8 S W r4c1 N E r4c2 N W r4c3 N S r4c4 E S r4c5 E W r4c6 N E W r4c7 W r4c8 N S r5c1 E S r5c2 S W r5c3 N S r5c4 N E r5c5 S W r5c6 S r5c7 S r5c8 N S r6c1 N S r6c2 N E r6c3 N S W r6c4 E S r6c5 N W r6c6 N S r6c7 N S r6c8 N S r7c1 N S r7c2 E r7c3 N W r7c4 N E r7c5 S W r7c6 N S r7c7 N E r7c8 N W r8c1 N E r8c2 E W r8c3 E W r8c4 E W r8c5 N E W r8c6 N E W r8c7 E W r8c8 W start r3c6 goal r8c2 [ r1c1 r8c8 r1c2 r8c7 r1c3 r8c6 ] R : r3c6 r8c2 r2c6 r8c3 r4c6 r8c1 r1c6 r8c4 r4c7 r7c1 r4c5 r8c5 r1c7 r6c1 r4c4 r7c5 r1c8 r8c6 r5c4 r5c1 r2c8 r7c4 r5c5 r7c6 r2c7 r8c7 r6c5 r6c4 A : r3c6 r4c6 r4c5 r4c4 r5c4 r5c5 r6c5 r6c4 r7c4 r7c5 r8c5 r8c4 r8c3 r8c2 \textless{}eos\textgreater{}\par}}

\par\medskip\noindent\textbf{S5: Dead-end filling}\par\smallskip
\noindent\fbox{\parbox{\dimexpr\linewidth-2\fboxsep-2\fboxrule\relax}{\footnotesize\raggedright\ttfamily Q : r1c1 E S r1c2 E W r1c3 E W r1c4 E S W r1c5 W r1c6 E S r1c7 E W r1c8 S W r2c1 N E r2c2 E S W r2c3 W r2c4 N E r2c5 S W r2c6 N S r2c7 E S r2c8 N W r3c1 S r3c2 N S r3c3 E S r3c4 E W r3c5 N W r3c6 N S r3c7 N E r3c8 S W r4c1 N E r4c2 N W r4c3 N S r4c4 E S r4c5 E W r4c6 N E W r4c7 W r4c8 N S r5c1 E S r5c2 S W r5c3 N S r5c4 N E r5c5 S W r5c6 S r5c7 S r5c8 N S r6c1 N S r6c2 N E r6c3 N S W r6c4 E S r6c5 N W r6c6 N S r6c7 N S r6c8 N S r7c1 N S r7c2 E r7c3 N W r7c4 N E r7c5 S W r7c6 N S r7c7 N E r7c8 N W r8c1 N E r8c2 E W r8c3 E W r8c4 E W r8c5 N E W r8c6 N E W r8c7 E W r8c8 W start r3c6 goal r8c2 [ r8c1 r8c2 r8c3 r1c8 r2c8 r3c8 ] R : r1c5 r2c3 r3c1 r4c1 r4c2 r4c7 r5c6 r5c7 r6c6 r6c7 r7c2 r7c3 r7c6 r7c7 r7c8 r8c8 r3c2 r6c8 r8c7 r2c2 r5c8 r8c6 r2c1 r4c8 r1c1 r1c2 r1c3 r1c4 r2c4 r2c5 r3c5 r3c8 r3c4 r3c7 r2c7 r2c8 r3c3 r4c3 r5c3 r6c3 r1c8 r6c2 r1c7 r5c2 r1c6 r2c6 r5c1 r6c1 r7c1 r8c1 A : r3c6 r4c6 r4c5 r4c4 r5c4 r5c5 r6c5 r6c4 r7c4 r7c5 r8c5 r8c4 r8c3 r8c2 \textless{}eos\textgreater{}\par}}

\subsection{Math}
\begin{figure}[t!]
    \centering
    \includegraphics[width=\linewidth]{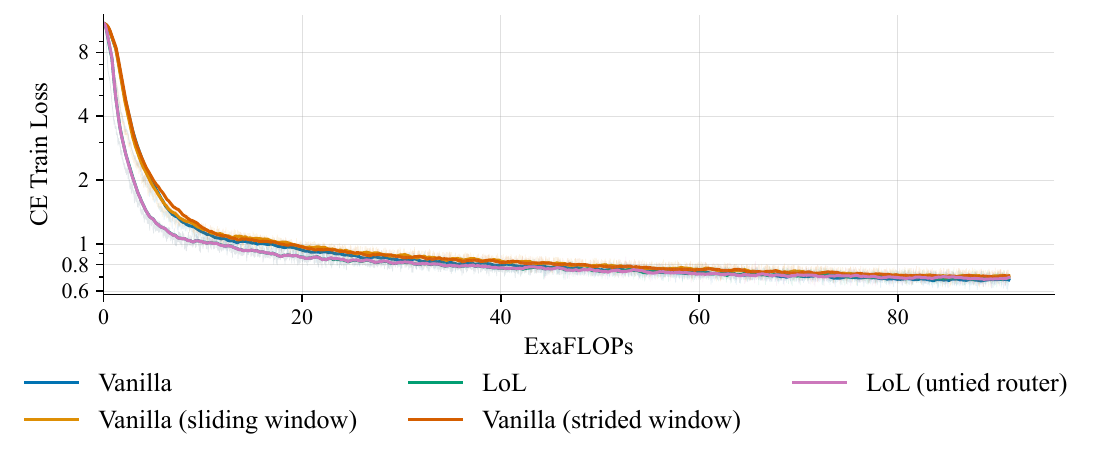}
    \caption{Extended version of \Cref{fig:math} with ``Vanilla (strided window)'' and ``LoL (untied router).''}
    \label{app-fig:math}
\end{figure}

In \Cref{app-fig:math} and \Cref{app-tab:math}, we extend \Cref{fig:math} and \Cref{tab:math} with ``Vanilla (strided window)'' and ``LoL (untied router).''
For our evaluations, we use \(5\) shot evaluation, except for GSM8K, where we use \(3\) shot evaluation to fit within the \(1024\) token context.

\begin{table}[t!]
\centering
\small
\begin{tabular}{lccccc}
\toprule
Model & ARC-E & ARC-C & HellaSwag & PIQA & GSM8K \\
\midrule
Random & 25.0 & 25.0 & 25.0 & 50.0 & 0.0 \\
\midrule
Vanilla & $40.8 \pm 1.0$ & $28.2 \pm 1.3$ & $31.4 \pm 0.5$ & $56.6 \pm 1.2$ & $29.7 \pm 1.3$ \\
Vanilla (sliding window) & $49.5 \pm 1.0$ & $28.5 \pm 1.3$ & $31.3 \pm 0.5$ & $58.8 \pm 1.1$ & $27.5 \pm 1.2$ \\
\name & $50.3 \pm 1.0$ & $29.1 \pm 1.3$ & $31.2 \pm 0.5$ & $59.7 \pm 1.1$ & $27.9 \pm 1.2$ \\
Vanilla (strided window) & $48.5 \pm 1.0$ & $28.0 \pm 1.3$ & $30.8 \pm 0.5$ & $59.0 \pm 1.1$ & $25.9 \pm 1.2$ \\
\name (untied router) & $50.1 \pm 1.0$ & $30.0 \pm 1.3$ & $31.4 \pm 0.5$ & $57.1 \pm 1.2$ & $27.7 \pm 1.2$ \\
\bottomrule
\end{tabular}
\caption{Extended version of \Cref{tab:math} with ``Vanilla (strided window)'' and ``LoL (untied router).''}
\label{app-tab:math}
\end{table}

\subsection{Looped Transformers}
\begin{figure}[ht!]
    \centering
    \includegraphics[width=1\linewidth]{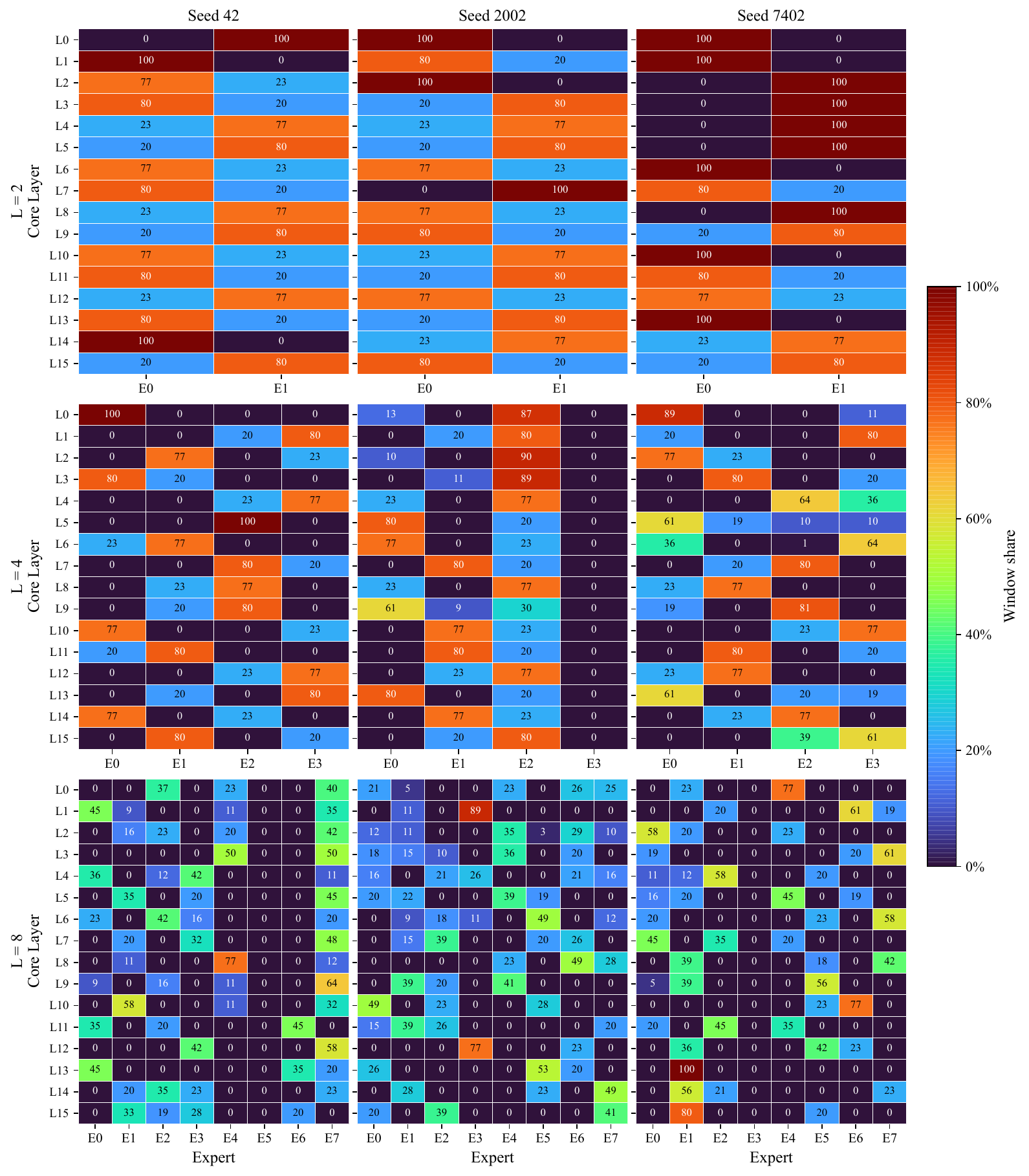}
    \caption{Heatmap of expert utilization of \name models shown in \Cref{fig:huginn}.}
    \label{app-fig:looped_seeds}
\end{figure}

In \Cref{app-fig:looped_seeds}, we visualize the expert utilization of the models shown in \Cref{fig:huginn}.

\subsection{MoE}
\begin{figure}[ht!]
    \centering
    \includegraphics[width=0.6\linewidth]{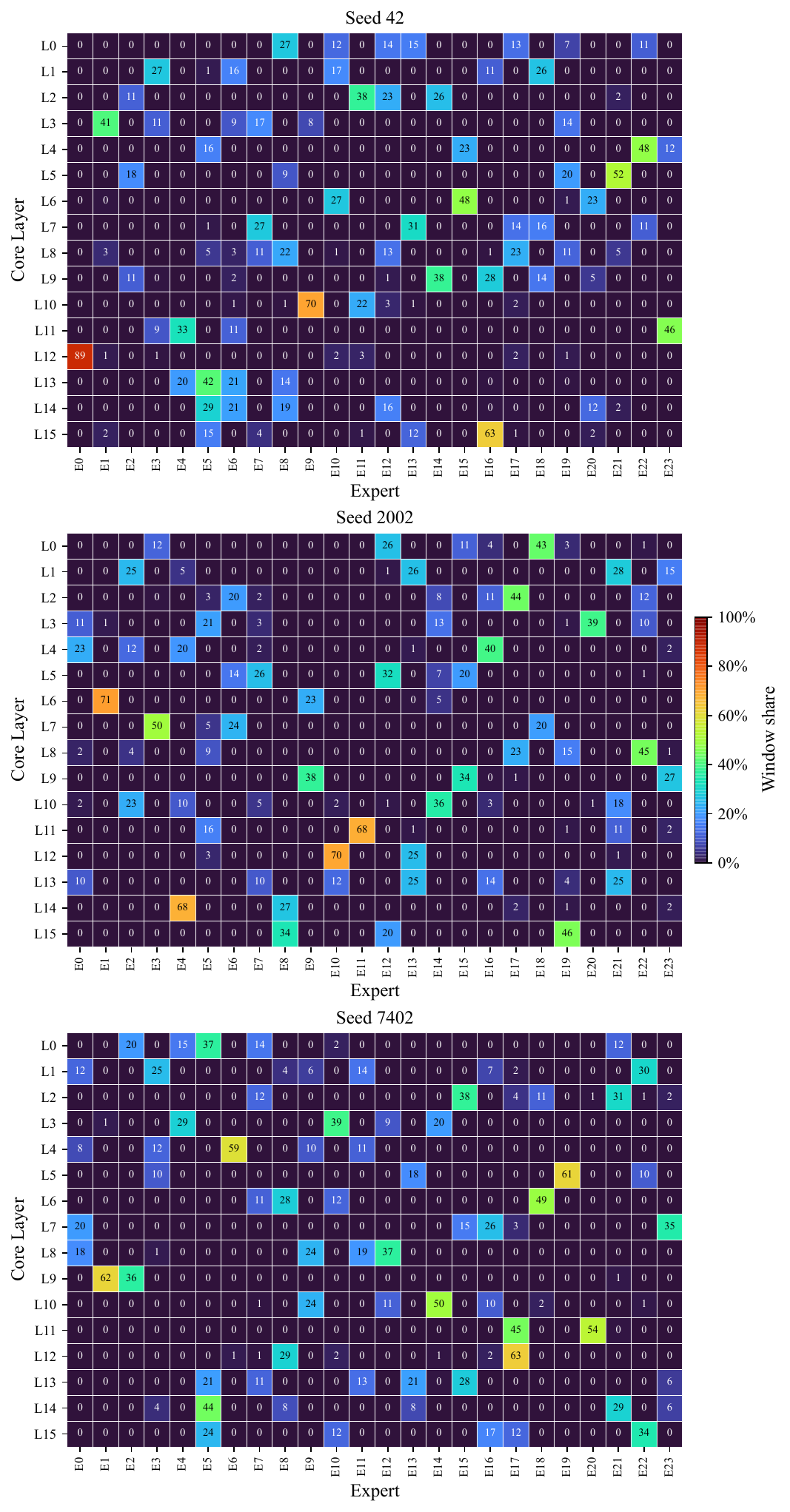}
    \caption{Heatmap of expert utilization of \name models shown in \Cref{fig:moe}.}
    \label{app-fig:moe_seeds}
\end{figure}

In \Cref{app-fig:moe_seeds}, we visualize the expert utilization of the models shown in \Cref{fig:moe}.

\end{document}